%% file: cameraready.tex
\documentclass{article} 
\usepackage{iclr2025_conference,times}

\input{math_commands.tex}

\usepackage{hyperref}
\usepackage{url}
\usepackage{graphicx}
\usepackage{amsmath}
\usepackage{algorithm2e}
\usepackage{booktabs}

\title{Intrinsic Dimension Correlation:\\ uncovering nonlinear connections in\\ multimodal representations}

\author{Lorenzo Basile $^{1,2}$ \quad Santiago Acevedo $^{3}$ \quad Luca Bortolussi $^1$ 
\\
\textbf{Fabio Anselmi} $^{1,4}$ \quad \textbf{Alex Rodriguez} $^{1,5}$
\vspace{1mm}\\
    $^1$University of Trieste \quad 
    $^2$AREA Science Park \quad
    $^3$SISSA \quad
    $^4$MIT \quad
    $^5$ICTP\\
\texttt{lorenzo.basile@phd.units.it}
\vspace{-3mm}
}

\newcommand\freefootnote[1]{%
  \let\thefootnote\relax%
  \footnotetext{#1}
  \let\thefootnote\svthefootnote
}

\iclrfinalcopy 
\begin{document}

\maketitle
\begin{abstract}
\looseness=-1 To gain insight into the mechanisms behind machine learning methods, it is crucial to establish connections among the features describing data points. However, these correlations often exhibit a high-dimensional and strongly nonlinear nature, which makes them challenging to detect using standard methods. 
 This paper exploits the entanglement between intrinsic dimensionality and correlation to propose a metric that quantifies the (potentially nonlinear) correlation between high-dimensional manifolds. We first validate our method on synthetic data in controlled environments, showcasing its advantages and drawbacks compared to existing techniques. Subsequently, we extend our analysis to large-scale applications in neural network representations. Specifically, we focus on latent representations of multimodal data, uncovering clear correlations between paired visual and textual embeddings, whereas existing methods struggle significantly in detecting similarity. Our results indicate the presence of highly nonlinear correlation patterns between latent manifolds.
\end{abstract}

\section{Introduction}

\begin{figure}[t!]
  \centering
  \includegraphics[width=1\textwidth]{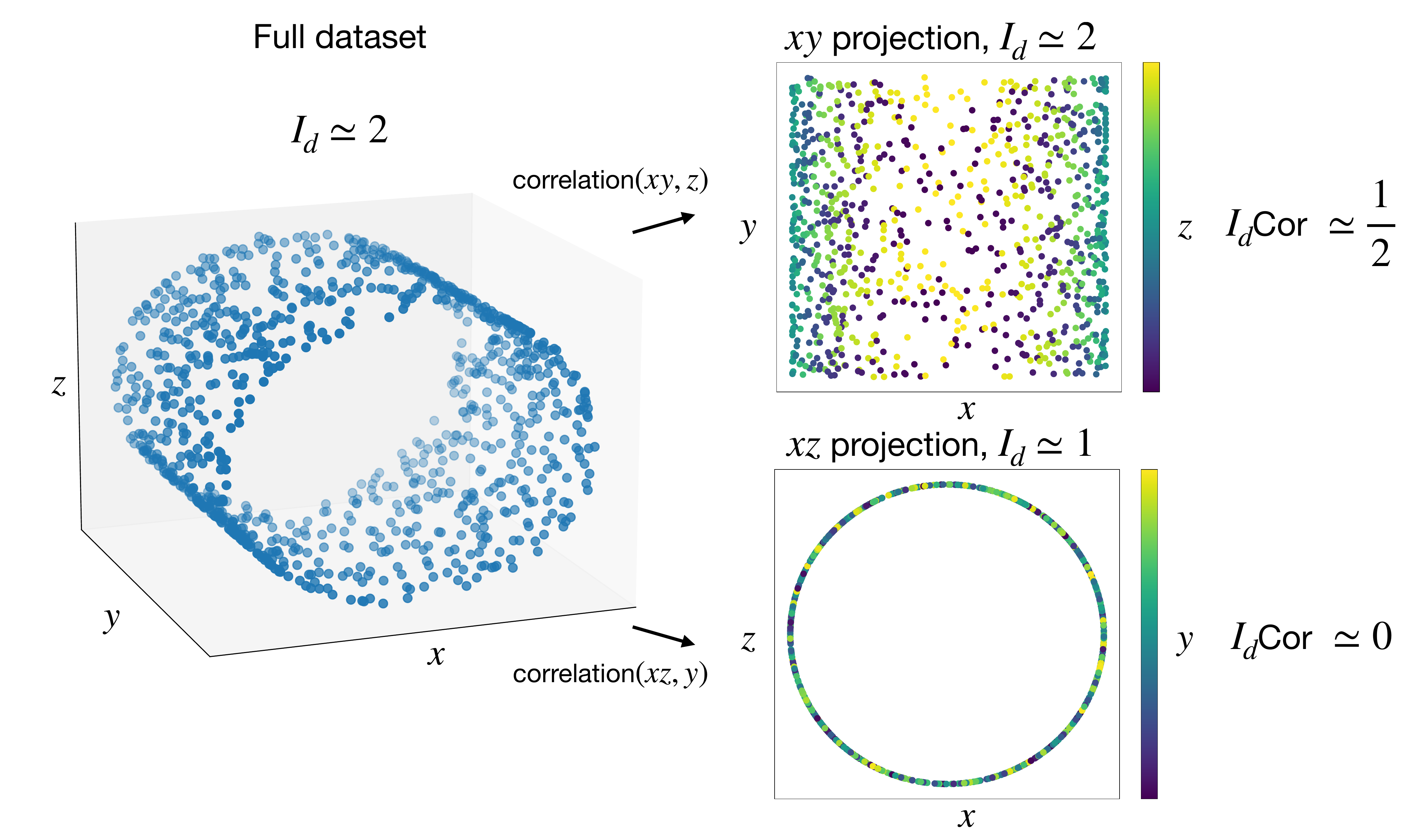}
  \caption{
  \looseness=-1 Example usage of $I_d$Cor: we consider a $3$D dataset in which the points lie on the surface of a cylinder, hence whose intrinsic dimension ($I_d$) is $2$.  We want to assess the correlation between the $2$D set of coordinates $xy$ (which also has $I_d=2$, as shown in the top-right panel) and the $1$D set $z$. Intuitively, {as $x$ and $z$ describe a circle}, it is evident that knowledge of $z$ (encoded by color) is very informative in determining the $x$ coordinate (e.g., a yellow point is sure to be found in the central region of the $x$ axis), but not in determining the $y$: hence, the correlation coefficient is $0.5$, according to equation \ref{idcor_eq}. Conversely, when estimating the correlation between $xz$ (whose $I_d$ is $1$, bottom-right panel) and $y$, we can see that having access to $y$ (which is now represented by color) does not give any information on the value of $x$ nor $z$, hence the correlation is $0$. }\label{cylinder}
\end{figure}

 Modern machine learning models have the remarkable ability to extract subtle patterns from complex datasets and use them to perform a wide variety of tasks in an astonishingly accurate way. However, to date, we still lack a complete and accurate understanding of their inner workings, especially in the case of deep neural networks.
An active field of research in the interpretability of neural networks is focused on characterizing and quantifying the similarity between different models. To this aim, many works \citep{raghu2017svcca, kornblith2019similarity, nguyen2021do} evaluate the statistical correlation between the latent representations produced by the models.
This quantification is key because it allows, for example, to disentangle or tie together different aspects of data representations, allowing a better interpretation of how the model makes its decisions. Moreover, assessing the similarity between representations of different models is particularly useful to determine whether the latent spaces are \textit{compatible}, meaning that information extracted by one model can be successfully transferred to  others. 

This point is crucial particularly when data points are represented by multiple interrelated modalities, such as visual and textual. In recent times, it has been shown \citep{norelli2023asif, moayeri2023text} that it is possible to build multimodal vision-language models starting from pre-trained unimodal encoders, in an effort to match the outstanding performance of state-of-the-art vision-language models such as CLIP \citep{radford2021learning}. These findings indicate that modern deep models can produce compatible representations when evaluated on aligned text-image data, hinting at a strong functional similarity between those. However, we find that standard latent similarity metrics such as CKA \citep{kornblith2019similarity} and Distance Correlation \citep{szekely2007measuring, zhen2022versatile} find a very low structural correlation between paired multimodal representations. We hypothesize that this is due to the strongly nonlinear nature of these correlations, making them difficult to analyze with standard methods. 

Our work tackles this challenge by introducing a novel metric, dubbed Intrinsic Dimension Correlation ($I_d$Cor) that leverages the concept of intrinsic dimension ($I_d$), i.e., the minimum number of variables required to describe the data, to quantify the mutual information between high-dimensional data manifolds. 
 Intuitively, the metric is based on the concept that if two data representations are correlated, the intrinsic dimension of a dataset created by concatenating the features of these representations is reduced, because the information in one representation can describe some aspects of the other. An example is provided in Fig. \ref{cylinder}. It is worth noting that, to compute the correlation between two representations, a simpler approach based on the difference between the embedding dimension and the intrinsic dimension of the concatenated dataset would not suffice, since this would measure the amount of correlated features, regardless of whether these correlations are \textit{intra} or \textit{inter} representation. Computational experiments indicate that our method is effective at detecting nonlinear relationships where traditional methods often fail.

Thus, the main contributions of this work can be summarized as follows:
\begin{itemize}
    \item  We propose a novel perspective that links the concepts of correlation and intrinsic dimension. The key idea is that, since the $I_d$ serves as a proxy for the information content of a dataset, the $I_d$ of a dataset created by merging two datasets will represent their joint information.
    \item Building on this idea, we propose $I_d$Cor, a novel correlation metric based on intrinsic dimension estimation, able to unveil nonlinear correlation between high-dimensional data manifolds, even of unpaired dimensions. We evaluate our metric on synthetic data, showcasing its strengths and limitations in comparison with existing methods.
    \item We then consider more complex scenarios and quantify the correlation between the representations learned by different deep neural networks engaged in large datasets. Specifically, we focus on multimodal data, demonstrating that we can find strong evidence of correlation where standard methods struggle to identify any. 
\end{itemize}

\section{Background}

\subsection{Correlation in latent representations}
Understanding whether different neural networks can learn to process data in \textit{similar} ways is a crucial point when trying to make sense of their results. 
Despite the inherent difficulty in defining what it means for two neural networks to be similar (or, at least, to behave similarly), research in this field has made significant progress in the last few years. A well-established approach to test if two neural networks are similar is that of measuring the statistical correlation between the data representations (or embeddings) learned by them. Recent proposals in this direction include Singular Value Canonical Correlation Analysis (SVCCA) \citep{raghu2017svcca}, Projection Weighted CCA (PWCCA) \citep{morcos2018insights}, Centered Kernel Alignment (CKA) \citep{kornblith2019similarity}, Distance Correlation (dCor) \citep{szekely2007measuring, zhen2022versatile}, Aligned Cosine Similarity \citep{hamilton2016diachronic} and Representation Topology Divergence (RTD) \citep{barannikov2022representation}, among others. For a more complete summary of current approaches to neural network similarity measurement, we defer the reader to \citet{klabunde2023similarity}.

These techniques have been widely employed to gain a deeper understanding of various aspects of the way neural models process information: for instance to quantify how different vision architectures encode spatial information \citep{raghu2021vision, nguyen2021do} or the relation between learning disentangled features and adversarial robustness \citep{zhen2022versatile}. Taking a slightly different approach, \citet{davari2023reliability} provides an in-depth analysis of the sensitivity of CKA to transformations that occur frequently in neural latent spaces, showcasing the importance of gathering results from a broader range of similarity metrics to obtain reliable information.

\subsection{Multimodal latent space alignment}
An example that highlights the importance of assessing similarity, quantified via correlation measures, between neural representations is further demonstrated by the recent empirical findings related to the so-called \textit{latent communication}. This concept, introduced by \citet{moschella2023relative}, builds on the idea that it is possible to transfer knowledge between latent spaces, even when they are produced by different models and on different data modalities, provided that some semantic alignment between the data exists (for example, images and their textual descriptions). The feasibility of this knowledge transfer was shown in \citet{moschella2023relative} through the introduction of \textit{relative} representations, where each point of the original representation is mapped according to its distance from a set of fixed \textit{anchor} points. Using this alternative representation of data, the authors show that it is possible to \textit{stitch} \citep{lenc2015understanding} together encoders and decoders coming from different models, with little to no additional training. 

Furthermore, numerous recent studies have demonstrated that large state-of-the-art visual and textual encoders can produce transferable representations when evaluated on aligned data (i.e., the same data or data that share some semantics, such as image-caption pairs). Indeed, a simple linear transformation is usually enough to map one latent space into another \citep{moayeri2023text, merullo2023linearly, maiorca2023translation, lahner2023direct}, at least in terms of performance on a specific downstream task, e.g., classification. 
It is worth noting that, to perform the alignment of the data, one assumes prior knowledge about the semantic correlation between the data representations (in order to define the anchor points). Hence, while these findings suggest a remarkable similarity between compatible latent spaces, the problem of detecting these correlations without relying on any downstream evaluation is still an open problem. Indeed, our investigation into the connection between aligned textual and visual embeddings reveals a very weak correlation using existing methods, calling for the development of methods that allow the identification of nonlinear correlations in high-dimensional spaces.

\subsection{Intrinsic dimension}

The concept of the intrinsic dimension ($I_d$) of a dataset is widely used in data analysis and Machine Learning. Before providing a more formal definition, imagine a dataset where your data points are the cities around the globe described by their 3D Cartesian coordinates. We will say that the {\it embedding dimension} of this dataset is three. However, anyone familiar with cartography would agree that nearly the same information can be encoded with only two coordinates (latitude and longitude). Therefore, its $I_d$ would be equal to two. Indeed, one of the definitions of $I_d$ is the minimum number of coordinates needed to represent the data with minimal information loss. A complementary definition is the dimension of the manifold in which the data lie, which in this case would be a sphere.

A possible way of estimating the $I_d$ is to find a meaningful projection  (i.e., with minimal information loss) into the lowest dimensional space possible. 
A classical method for doing that is Principal Component Analysis \citep{wold1987principal}, but it has the drawback that the intrinsic dimension it estimates is only correct if the manifold in which the data lie is a hyperplane.
Therefore, the development of methods that can estimate the $I_d$ in nonlinear manifolds is an active research field. Typically, these approaches infer the $I_d$ from the properties of distances to the Nearest Neighbors. While a full review of these methods is out of the scope of this work (the interested reader is referred to \citet{campadelli2015intrinsic}), it is worth mentioning the Maximum Likelihood estimator (MLE)~\citep{levina2004maximum}, the Dimensionality from Angle and Norm Concentration (DANCo) approach~\citep{ceruti2014danco} or the TwoNN \citep{facco2017estimating}. The last is the one employed in this work since it is particularly fast and behaves well even in the case of datasets with a high non-uniformity on the density of points. A brief description of TwoNN is provided in Appendix \ref{twonn_appendix}.

More recently, several studies have estimated the intrinsic dimension of neural representations, demonstrating that \( I_d \) is a valuable tool for understanding the geometry of the latent manifolds produced by deep models. This concept was initially explored in \citet{ansuini2019intrinsic}, where the authors estimated \( I_d \) across different layers of CNNs, gaining insights into the sequential information flow within these models. Later, \citet{valeriani2023geometry} and \citet{cheng2023bridging} analyzed the representations of transformer models, across different domains, while \citet{kvinge2023exploring} studied the internal $I_d$ of generative diffusion models.
In a slightly different direction, \citet{brown2022relating} unveiled a connection between generalization and the $I_d$ of the hidden representations, while \citet{kaufman2023data} studied the relation between $I_d$ and curvature in latent manifolds.

\section{Correlation through Intrinsic Dimension}\label{idcor}

The intrinsic dimension of a dataset is closely linked to the correlations among the various features that define the data points. These correlations determine the regions in which the data points can exist, thereby shaping the underlying manifold. 
Let us consider the simplest example: a two-dimensional dataset. If the two variables are uncorrelated, their linear correlation coefficient ($R^2$) approaches zero while, if one feature is a linear function of the other, $R^2$ becomes equal to one. The two scenarios differ by the $I_d$ of the data manifold: the first case corresponds to a plane ($I_d = 2$), while the second corresponds to a line ($I_d = 1$). If we examine a slightly more complex case, the advantage of exploiting the $I_d$ for correlation becomes evident. Let us consider a spiral-shaped dataset embedded in two dimensions: it has $R^2 \approx 0$ due to the nonlinear nature of the correlation between the two variables, while the behavior of the $I_d$ is identical to the one observed on the linearly correlated dataset, as reported in detail in section \ref{2dexp}. 

A formal framework that connects correlation and intrinsic dimension comes from information theory \citep{romano2016measuring}. In this field, mutual information is the fundamental metric that quantifies the relationship between simultaneously sampled random variables. Mutual information makes a natural candidate to serve as a measure of correlation between data, and \citet{horibe1985entropy} defined a correlation coefficient between two random variables $X$ and $Y$ (known as Normalized Mutual Information - NMI) as:
\begin{equation}\label{horibe}
    \varrho = \frac{I(X,Y)}{\max(H(X), H(Y))}
\end{equation}
where $I(\cdot, \cdot)$ is the mutual information and $H(\cdot)$ is the entropy. 

In high dimensional datasets, computing the mutual information is challenging due to the curse of dimensionality, so we introduce the intrinsic dimension as a proxy.
Mutual information can be expressed in terms of entropies as:
\begin{equation}
    I(X,Y)=H(X)+H(Y)-H(X,Y)
\end{equation}
Recent research \citep{bailey2022local, ghosh2023local} has shown that intrinsic dimension is a metric that can satisfy the most desirable properties of entropy, while being easy and fast to compute on large-scale data (in Appendix \ref{example_id}, we provide a test case where this relationship is exact). We exploit this relation to define our correlation coefficient $I_d$Cor, based on the NMI formulation, but using intrinsic dimension as a replacement for entropy. Given two standardized datasets (of possibly different dimensionality) $X \in \mathbb{R}^{n,d_1}$ and $Y\in \mathbb{R}^{n,d_2}$, we define their $I_d$Cor as:

    \begin{equation}\label{idcor_eq}
    I_{d}\text{Cor}(X,Y)=\frac{(I_d(X)+I_d(Y)-I_d(X\oplus Y))}{\max(I_d(X),I_d(Y))}
\end{equation}
where $\oplus$ denotes row-wise concatenation.
Although various other versions of NMI have been proposed over time (for a complete survey on variations of NMI we defer the reader to \citet{vinh2009information}) we adopt the $\max$ normalization since it allows an interpretation as the fraction of features of the most informative dataset that can be predicted using the less informative one (see Fig. \ref{cylinder} for an intuitive explanation of this concept).

In practical applications, intrinsic dimension is calculated using estimators, which can be prone to errors. To mitigate this, we assign a $p$-value to the observed correlation, employing a permutation test \citep{davison1997bootstrap} on $I_d(X\oplus Y)$.
Specifically, we estimate the $I_d$ of several independent samples of the joint dataset, created by
concatenating the two original datasets and randomizing the pairings to disrupt any existing correlations.
This process allows us to determine a $p$-value that represents the probability of the hypothesis of the two datasets being uncorrelated as $p=\frac{L+1}{S+1}$, where $L$ is the number of estimates lower than the original joint $I_d$ and $S$ is the total number of permuted samples considered.

\section{Results}
In this section, we begin by assessing our proposed $I_d$Cor measure in simple, controlled settings using synthetic data. Then, we move to applications to latent representations generated by different neural networks on various datasets. We begin with a straightforward example that underscores the difficulty faced by standard methods in identifying nonlinear relationships, then progress to more extensive applications involving visual representations and multimodal text-image representations.
For some experiments, we do not report the correlation $p$-value returned by our method. In such cases, the $p$-value is always the lowest possible according to the permutation test outlined in section~\ref{idcor} ($\frac{1}{S+1}$, where $S$ is the number of permutations, $100$ in most of our experiments).

\subsection{Synthetic experiments}\label{2dexp}
As a first step, we produce three toy datasets, displayed in the Appendix in Fig. \ref{toyexample}. Such datasets are made of 5000 observations of two variables that are either linearly correlated, uncorrelated, or nonlinearly correlated. The first setting is simply obtained by arranging $x$ and $y$ on a straight line, in the second case both random variables are sampled independently from a normal Gaussian distribution, while the last dataset contains data arranged on a spiral curve.

We report our correlation results in Table \ref{toy_results}, aggregating results over 10 independent random samplings of the datasets. In the simpler cases (linear and random data) our method agrees with linear correlation and distance correlation, correctly identifying a very strong correlation in linear data and the lack thereof in random data. The spiral dataset constitutes a more tricky testbed: while there is a clear correlation between the two variables, it is a highly nonlinear one, and the linear correlation coefficient is around $0$. Even Distance Correlation, despite being a nonlinear method, fails to find any strong signal of correlation, returning a value very close to $0$. Instead, our method correctly identifies the strong dependency between the two variables, with a mean correlation coefficient of $0.98$, determined with high confidence, as witnessed by the $p$-value consistently equal to $0.01$. 
We note here that our method relies on an intrinsic dimension estimator (in this paper, we mainly employ TwoNN \citep{facco2017estimating}, but we also provide results on synthetic data using MLE \citep{levina2004maximum} in section \ref{mle_appendix} in the Appendix), and it inherits substantial properties from it. On the negative side, $I_d$ estimators are not oracles, and they can return values that slightly differ from what one would expect (e.g., $I_d$ lower than $1$ in our linear dataset or higher than $2$ in the random case), or even totally fail (when the $I_d$ becomes large enough, the estimator is also affected by the curse of dimensionality). Conversely, the choice of employing TwoNN makes our method extremely efficient (the correlation coefficient can be obtained with just 3 calls to the estimator), allowing it to scale easily to large and high-dimensional datasets.

\begin{table}[h]
\centering 
  \caption{Correlation in two-variable datasets.
  ($\mathbf{R^2}$): linear correlation coefficient;
  ($\mathbf{dCor}$): distance correlation coefficient; ({$\mathbf{I_d \oplus}$}): intrinsic dimension of the concatenated (2D) dataset; ($\mathbf{I_dCor}$): correlation coefficient ($\varrho$) returned by our method; ($\mathbf{p}$\textbf{-value}): significance of the correlation detected by our method ($100$ shuffles). Results are reported as $\text{mean}\pm\text{std}$ over 10 independent random samplings of data, with the exclusion of $p$-value, which is reported as a range.\\}
  \label{toy_results}
  \centering
  \begin{tabular}{cccccc}
    \toprule
    \textbf{Data} & $\mathbf{R^2}$    & $\mathbf{dCor}$&  $\mathbf{I_d \oplus}$ & $\mathbf{I_dCor}$ & $\mathbf{p}$\textbf{-value}\\
    \midrule
    Linear & $1.00\pm0.00$ &  $1.00\pm0.00$ & $0.99\pm0.02$ & $1.00\pm0.00$ & $0.01$  \\
    Random & $0.00\pm0.00$ &  $0.00\pm0.00$ & $2.02\pm0.04$ & $-0.02\pm0.04$ & $[0.03, 0.99]$  \\
    Spiral & $0.01\pm0.00$ &  $0.02\pm0.00$ & $1.01\pm0.02$ & $0.98\pm0.02$ & $0.01$  \\
    \bottomrule
  \end{tabular}
\end{table}

Then, we switch to a higher-dimensional setting, in which we test our method on correlating pairs of synthetic datasets of 4 variables. We construct a first random dataset containing 4 independently sampled variables, and then consider 3 correlation scenarios: in the first, we compute its correlation with another likewise randomly sampled dataset, hence we expect no correlation; in the second, we randomly sample 2 variables of the second dataset, while binding the other 2 to the original dataset through trigonometric functions, establishing a partial correlation between the two; finally, in the third scenario we set all variables of the second dataset to be trigonometric functions of the original variables, thus making the datasets completely correlated, although in a strongly nonlinear fashion.
We compare our $I_d$Cor against Distance Correlation and Canonical Correlation Analysis (CCA)~\citep{hotelling1936relations}, which is the direct analogous of $R^2$ in multivariate correlation, and serves as a linear baseline. As we report in Table \ref{4d_results}, while all methods agree in correctly finding no correlation in the first scenario, both CCA and dCor fail in the partially and completely correlated cases, while $I_d$Cor returns substantially higher correlation coefficients, showcasing its robustness to nonlinear correlations. In the Appendix (sections \ref{uniform_appendix} and \ref{noise_appendix}), we report additional results in this setting, obtained applying $I_d$Cor to differently sampled and noisy data.

\begin{table}[h]
\centering 
  \caption{Correlation between high dimensional synthetic datasets in various scenarios. We consider a first dataset with $5000$ observations of $4$ variables $w$, $x$, $y$ and $z$, all sampled independently from a Gaussian distribution with mean $0$ and standard deviation $\pi$, and assess correlations in 3 different scenarios. In scenario \textbf{A - absence of correlation} we compute the correlation between such dataset and another dataset built by independently sampling corresponding $w'$, $x'$, $y'$ and $z'$ from the same distribution. In scenario \textbf{B - partial correlation}, we randomly sample $y'$ and $z'$, while we set $w'=\cos(w+y)$ and $x'=\sin(x)$. Finally, in scenario \textbf{C - complete correlation}, we set $w'=\cos(w+y)$, $x'=\sin(x)$, $y'=\sin(xz)$, $z'=\cos(y)$. 
  Results are reported as $\text{mean}\pm\text{std}$ over 10 independent random samplings of data, with the exclusion of $p$-value, reported as a range.\\}
  \label{4d_results}
  \centering
  \begin{tabular}{cccccccc}
    \toprule
    \textbf{Correlation} & $\mathbf{CCA}$ & $\mathbf{dCor}$ &  $\mathbf{I_d \oplus}$& $\mathbf{I_d Cor}$ & $\mathbf{p}$\textbf{-value}\\
    \midrule
    (A) Absent & $0.02\pm0.01$ & $0.00\pm0.00$ & $8.07\pm0.17$ & $0.01\pm0.05$ & $[0.25, 1.00]$ \\
    (B) Partial & $0.02\pm0.01$ & $0.00\pm0.00$ & $6.41\pm0.10$ & $0.35\pm0.03$ & $0.01$ \\
    (C) Complete & $0.02\pm0.00$ & $0.01\pm0.00$ & $4.88\pm0.06$ & $0.67\pm0.02$ & $0.01$ \\
    \bottomrule
  \end{tabular}
\end{table}

\subsection{A motivating example on neural representations}
\begin{figure}[h]
  \centering
  \includegraphics[width=0.9\textwidth]{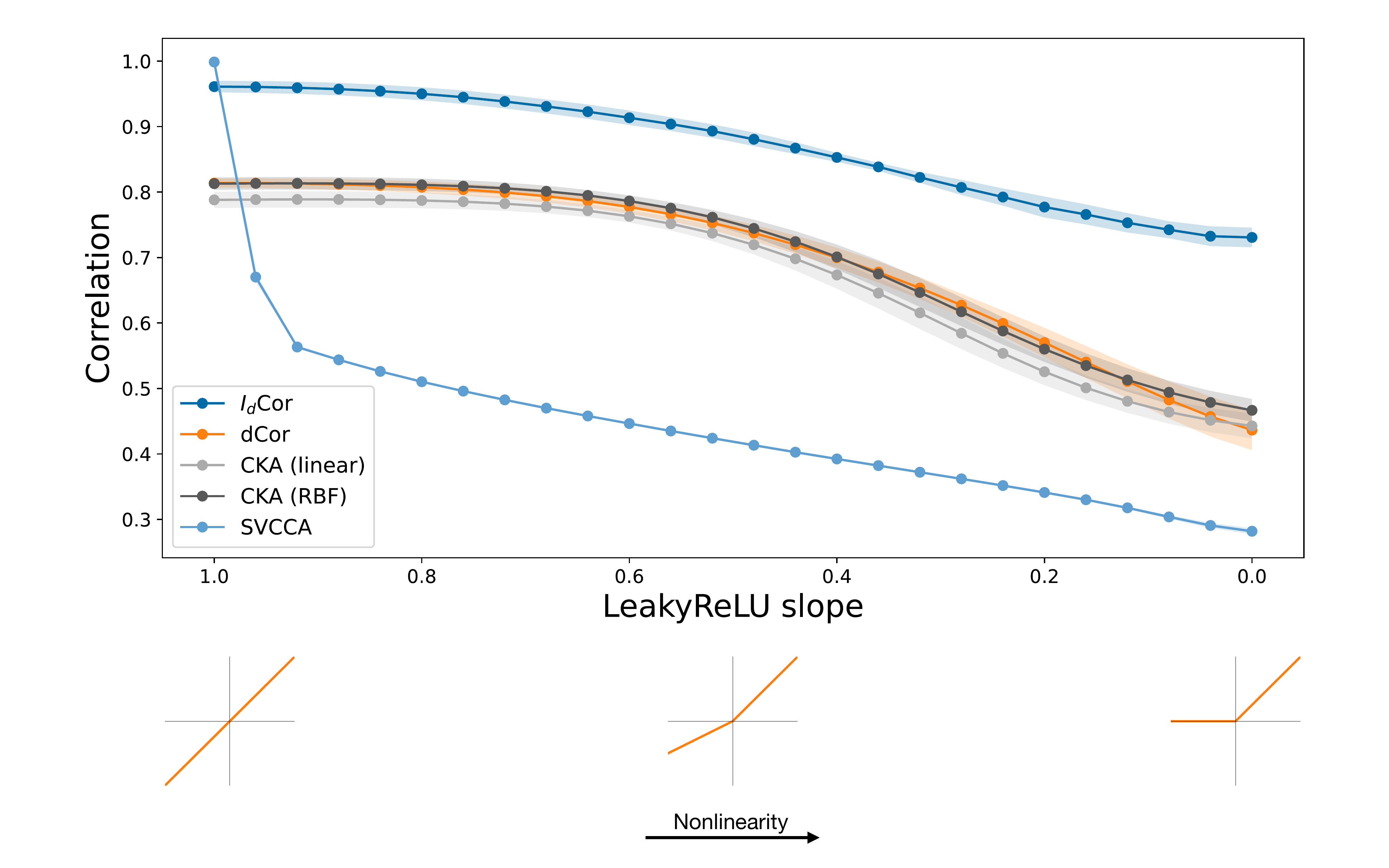}
  \caption{Average correlation results with different methods between MNIST data and their final representations computed by a randomly initialized MLP, with variable degree of activation nonlinearity, increasing on the $x$ axis. Shaded area represents standard deviation over 10 runs with independent random initialization of MLP weights.}\label{mnist}
\end{figure}
Across different layers, neural networks encode information in complex, high-dimensional representations that differ significantly from the simpler datasets discussed earlier in this manuscript. In particular, deep models are structured to learn nonlinear functions of the input data, typically through the use of nonlinear activation functions like ReLU. This suggests that the representations produced by different networks on the same data can be correlated in complex, nonlinear ways. Consequently, methods used to detect such correlations need to be capable of capturing this degree of nonlinearity.

To illustrate this phenomenon, we showcase a simple example: we consider a randomly initialized multilayer perceptron (MLP), made of $15$ fully connected layers of $784$ neurons, followed by a LeakyReLU activation. LeakyReLU is a parametric activation function, whose behavior depends on a parameter called slope: if the slope is $1$, it behaves like the identity function, rendering our MLP a linear function of the input, while lower slope values make the network nonlinear, with $0$ corresponding to the standard ReLU. We feed our MLP with the MNIST \citep{lecun1998gradient} dataset at increasing degrees of nonlinearity (which corresponds to decreasing the slope) and compute the correlation between the representation at the final layer and the input data, both with our method and with established baselines (SVCCA, Distance Correlation, linear kernel CKA, and RBF kernel CKA).

\looseness=-1 As we report in Fig. \ref{mnist}, our method is weakly affected by the increasing nonlinearity in the correlation, as it consistently returns correlation coefficients above $0.75$. Existing baselines capture high correlation in linear or quasi-linear cases (high slope), but the signal tends to degrade quickly as the slope decreases. This is especially true for SVCCA, as it is a linear method, but even nonlinear alternatives see the initial correlation fade when the activation becomes ReLU, reaching values below~$0.5$.

Our method estimates a proxy of the normalized mutual information between representations, being therefore largely insensitive to nonlinear transformations of the data (as opposed to other metrics like CKA or dCor) and making it optimal to detect statistical dependencies between representations. However, this insensitivity makes it a less informative geometrical similarity index, since the two representations will have a high mutual information regardless of the nature of the nonlinear transformation.

\subsection{ImageNet representations}\label{imagenet}

Moving to a more realistic setting, we test our method on measuring similarity between ImageNet \citep{russakovsky2015imagenet} embeddings coming from different neural encoders.
We consider a variety of pre-trained architectures, including CNNs (ResNet-18 \citep{he2016deep} and EfficientNet-B0 \citep{tan2019efficientnet}), four variants of Vision Transformers \citep{dosovitskiy2020image} (including a ViT-CNN hybrid) and three ViT-based self-supervised vision-language models (CLIP-ViT-B \citep{radford2021learning}, SigLIP-ViT-B \citep{zhai2023sigmoid} and BLIP-ViT-B \citep{li2022blip}). For all the models, we consider the output of the last hidden representation produced by the encoder, before the classification head: in supervised ViTs this choice corresponds to the last \textit{class} token, while in CNNs to the pooler output. For CLIP, SigLIP and BLIP, we consider instead the image representation in the shared vision-language space.
\begin{figure}[h]
  \centering
  \includegraphics[width=\textwidth]{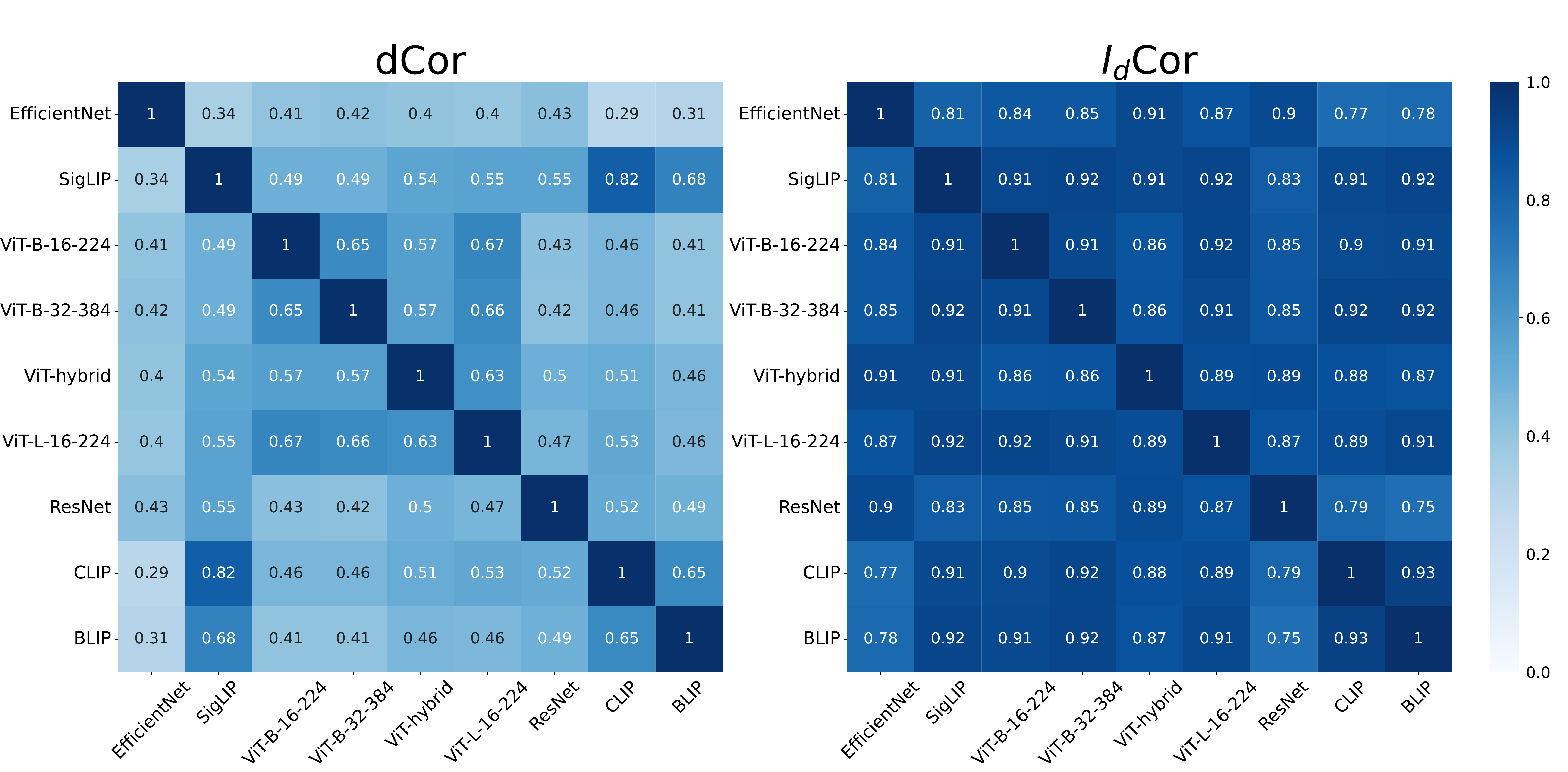}
  \caption{Correlation results on ImageNet representations, obtained using: \textbf{left} Distance Correlation (dCor); \textbf{right} $I_d$Cor (ours). Both methods are able to detect non-negligible correlation. More baseline results are reported in the Appendix.}\label{imagenetcorrelation}
\end{figure}
Since some of the correlation methods we employ as baselines (namely, CKA) are particularly expensive in terms of memory requirements, for all our experiments we randomly sample a subset of $30000$ data points.

We report the correlation results produced by our method in Fig. \ref{imagenetcorrelation}, along with the correlation scores returned by Distance Correlation (dCor), as it is the baseline method that most closely matches ours in terms of mean off-diagonal correlation ($I_d$Cor mean: $0.87$, dCor mean: $0.50$). Detailed results for SVCCA (mean: $0.46$), linear kernel CKA ($0.43$) and RBF kernel CKA ($0.44$) are reported in the Appendix in Fig. \ref{imagenetbaselines}.
In this setting, all models are computing embeddings for the same data points, hence we would expect significant correlation to be present for any given pair of models. Indeed, all methods are clearly capturing such correlation (even if with higher variance than $I_d$Cor), including SVCCA, which suggests that a relevant component of this correlation is actually linear.

\subsubsection{Coarse alignment}
The previous section demonstrated that our method effectively identifies strong correlations within perfectly aligned datasets of representations. We now aim to explore how the performance of the method might vary when applied to coarsely aligned data. To test this, we utilize the inherent class information of ImageNet data. Specifically, we randomly shuffle the embeddings generated by a model while keeping the labels unchanged, and then compare this modified dataset with the original dataset prior to shuffling. In other words, given a point index $i$ of class $C_i$, we pair it with another randomly chosen point $j$ of class $C_j$ with the condition that $C_i=C_j$.

This allows us to assess the robustness of our correlation estimation in less ideal conditions: we expect the correlation signal to decrease, as alignment is a crucial property for all correlation methods to detect similarities. However, as we report in Table~\ref{coarse}, we are still able to identify correlations with high confidence, as demonstrated by the low $p$-values. The values of the correlation suggest that the number of features needed to perform the classification task is between 50 and 75\%. On average, $I_d$Cor returns a correlation of $0.62$, while baselines sit in the range $0.31-0.47$. Interestingly, we observe that 
the models that exhibit lower correlation are CLIP, BLIP and SigLIP, all contrastive vision-language models. In two of the three cases, $I_d$Cor is even surpassed by Distance Correlation and CKA. While we have no clear explanation for such behavior, these results align with very recent results by \citet{ciernik2024training}, highlighting the impact of the pre-training objective on representation similarity.
For reference, we also report the $I_d$Cor results obtained when one of the two datasets is freely shuffled (irrespective of class labels). As witnessed by the high $p$-values, $I_d$Cor correctly reports a lack of correlation in such case.

\begin{table}[h]
\centering 
\scriptsize{
  \caption{Correlation between ImageNet representations when exact alignment is broken. Results are shown, in terms of $p$-value and correlation coefficient, for the coarse alignment case, in which data are shuffled while preserving the labeling. $I_d$Cor significantly outperforms all baseline methods, with the exception of two models. The last two columns report the $I_d$Cor coefficient and $p$-value for the fully shuffled case, in which no constraint is enforced on labels, and correlation is not present, as identified by high $p$-values. All results are reported as means (excluding $p$-values, reported as ranges) over $5$ independent shufflings. Standard deviations are not reported as they are lower than 0.02 in all cases.\\}
  \label{coarse}
  \centering
  \begin{tabular}{ccccccccc}
    \toprule
    Model & $I_d$Cor & $p$-val. & SVCCA & CKA (lin.) & CKA (RBF) & dCor& $I_d$Cor (rand) & $p$-val. (rand) \\
    \cmidrule(lr){1-1} \cmidrule(lr){2-7} \cmidrule(lr){8-9}
    EfficientNet & $\mathbf{0.61}$ & $0.01$ & $0.25$ & $0.17$ & $0.19$ & $0.25$ & $-0.10$ & $[0.16,0.97]$ \\
    SigLIP & $0.53$ & $0.01$ & $0.26$ & $0.58$ & $\mathbf{0.59}$ & $\mathbf{0.59}$ & $-0.01$ & $[0.09, 0.86]$ \\
    ViT-B-16 & $\mathbf{0.69}$ & $0.01$ & $0.39$ & $0.34$ & $0.34$ & $0.44$ & $0.27$ & $[0.27,0.98]$ \\
    ViT-B-32 & $\mathbf{0.68}$ & $0.01$ & $0.40$ & $0.40$ & $0.40$ & $0.47$ & $0.22$ & $[0.25,0.99]$ \\
    ViT-hyb. & $\mathbf{0.63}$ & $0.01$ & $0.39$ & $0.51$ & $0.52$ & $0.55$ & $0.27$ & $[0.09,0.97]$ \\
    ViT-L & $\mathbf{0.73}$ & $0.01$ & $0.40$ & $0.45$ & $0.43$ & $0.55$ & $0.37$ & $[0.18,0.91]$ \\
    ResNet & $\mathbf{0.66}$ & $0.01$ & $0.22$ & $0.34$ & $0.35$ & $0.37$ & $0.39$ & $[0.46,0.91]$ \\
    CLIP & $0.53$ & $0.01$ & $0.26$ & $0.59$ & $0.61$ & $\mathbf{0.62}$ & $0.21$ & $[0.08,0.85]$ \\
    BLIP & $\mathbf{0.50}$ & $0.01$ & $0.28$ & $0.37$ & $0.39$ & $0.39$ & $-0.10$ & $[0.28, 0.95]$ \\
    \cmidrule(lr){1-1} \cmidrule(lr){2-7} \cmidrule(lr){8-9}
    Average & $\mathbf{0.62}$ & $0.01$ & $0.31$ & $0.42$ & $0.42$ & $0.47$ & $0.17$ & $[0.08, 0.99]$\\
    \bottomrule
  \end{tabular}
  }
\end{table}

\subsection{Multimodal representations}\label{n24news}
We now shift our focus to a multimodal context, where we examine the similarities between hidden spaces learned by text and image encoders. We use three datasets, N24News \citep{wang2022n24news}, MS-COCO 2014 \citep{lin2014microsoft} and Flickr30k \citep{young2014image}, all consisting of image-caption pairs. This analysis will help us understand how textual and visual representations correlate when evaluated on related multimodal content. Images are encoded using a representative subset of the vision models introduced in section \ref{imagenet}: two CNNs (EfficientNet-B0 and ResNet-18), two ViTs (ViT-B-16 and ViT-hybrid), and the visual branch of CLIP.  For text we employ five architectures, all taken pre-trained: BERT \citep{devlin2019bert}, both cased and uncased, ALBERT \citep{lan2019albert}, Electra \citep{clark2020electra} and finally the text encoder of CLIP. For all text models, we consider the last representation of the \textit{class} token.

\begin{figure}[h]
  \centering
  \includegraphics[width=\textwidth]{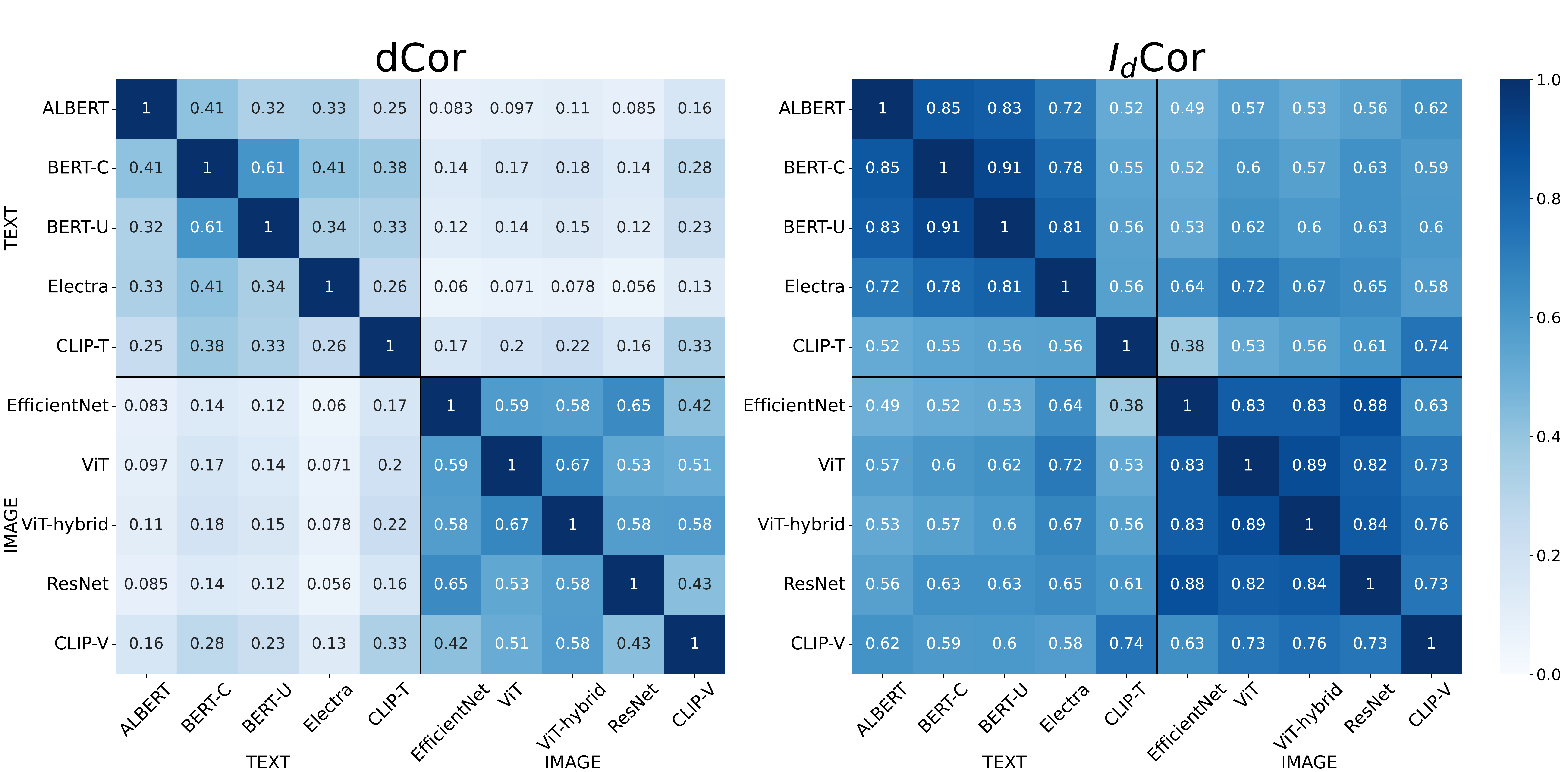}
  \caption{
  \looseness=-1 Correlation results on N24News representations, obtained using: \textbf{left} Distance Correlation (dCor); \textbf{right} $I_d$Cor (ours). Like all other baselines we evaluate, dCor is only able to spot correlations between encoders of the same modality, while $I_d$Cor reveals significant correlation for all model pairs.}\label{n24correlation}
\end{figure}

We report the correlation results obtained on N24News in Fig. \ref{n24correlation}, comparing our method against Distance Correlation (dCor), which is once again the closest-performing baseline method. Our $I_d$Cor yields a mean off-diagonal correlation of $0.66$, which is noticeably higher than those of baseline methods, in the range $(0.25-0.29)$.
In fact, the correlation heatmaps for all baseline methods reveal a clear block structure, as such methods are able to capture 
correlation only among same-modality encoders, but fail on cross-modal correlation. Full results for baseline methods are available in the Appendix (Fig. \ref{n24baselines}). Instead, our method returns significant correlation values even across different modalities, in accordance with previous findings \citep{maiorca2023translation} that proved N24News representations to be transferable across models and modalities.
Experiments on Flickr30k and MS-COCO confirm the behavior observed for N24News, as discussed in the Appendix (section \ref{flickrcoco}). Moreover, in sections \ref{5captionsappendix} and \ref{sanityappendix}, we provide additional experiments on the discussed datasets to show the discriminative power of $I_d$Cor in cases with low or no correlation, while section \ref{rtd} contains additional similarity results on N24News, obtained using RTD \citep{barannikov2022representation}.

\subsection{Computational resources}\label{resources}
We performed all the computations on a NVIDIA A100 GPU, equipped with 40GB of RAM. The main computational hurdle of our method is the estimation of $I_d$ through TwoNN: our implementation follows closely that of \citet{ansuini2019intrinsic}, which we translated to PyTorch to enable GPU acceleration. {With this setup, $I_d$Cor runs in the order of $1$s on two $1024$-dimensional datasets of $30000$ points}. Just like our correlation method, all the representation similarity baselines we employ greatly benefit from GPU acceleration: we used them in their PyTorch implementations provided by \citet{miranda2021ultimate_anatome} (SVCCA), \citet{maiorca2024latentis} (CKA) and \citet{zhen2022versatile} (Distance Correlation), with minor adaptations. Pretrained models were obtained from the Transformers library by HuggingFace \citep{wolf2020transformers}, details on the checkpoints we employed are provided in the Appendix (section \ref{models}). Our code is available at \url{https://github.com/lorenzobasile/IDCorrelation}.

\section{Discussion}

Our work introduces Intrinsic Dimension Correlation ($I_d$Cor), a novel and robust method for detecting complex nonlinear correlations in high-dimensional spaces. Due to its flexibility, this method can be employed in a wide range of applications, from natural language processing to computer vision and beyond (including other fields of science, like physics), offering a new type of analysis to address how machine learning models represent data.
Remarkably, our results show the effectiveness of the method to detect a correlation signal in multimodal data where we know a correlation should exist but where standard methods struggle to identify any.
\paragraph{Limitations}\label{limitations} Among the possible drawbacks of the method, it is worth mentioning that it is fully dependent on the precision of the $I_d$ estimator, so, if the $I_d$ is wrongly predicted the method will fail to find correlations. This would likely happen when the $I_d$s involved are big, so even last-generation estimators will be affected by the curse of dimensionality.
\paragraph{Future directions} Our method lays the foundations 
for many interesting future research avenues. For example, it can be used to disentangle data representations by minimizing the correlation between representations from two or more datasets, through the concept of Total Correlation \citep{watanabe1960information}. This approach could be highly relevant in applications such as multimedia analysis, cross-modal retrieval, and data fusion, potentially resulting in more interpretable neural networks. 

In its present form, $I_d$Cor is only applicable to datasets with uniform $I_d$. However, some realistic datasets can present more than one manifold with different intrinsic dimensionalities. With the development of methods that allow estimating in a reliable way these local $I_d$s \citep{allegra2020data, dyballa2023ian}, we devise that $I_d$Cor can be applied to quantify the correlations locally. We provide a proof-of-concept example in the Appendix \ref{ian_appendix}. 

In conclusion, our method not only enhances the understanding of high-dimensional data correlations but also paves the way for innovative solutions in representation learning and interpretability, making it a valuable tool across many fields of machine learning.
\section*{Acknowledgements}
We acknowledge AREA Science Park for making the supercomputing platform ORFEO available for this work. We also thank Sanketh Vedula for insightful discussions on statistical testing. This work is supported by the PNRR project iNEST (Interconnected Nord-Est Innovation Ecosystem) funded by the European Union Next-GenerationEU (Piano Nazionale di Ripresa e Resilienza (PNRR) – Missione 4 Componente 2, Investimento 1.5 – D.D. 1058 23/06/2022, ECS\_00000043, CUP J43C22000320006)
\bibliography{references}
\bibliographystyle{iclr2025_conference}
\newpage
\appendix
\section{Appendix}
\subsection{The TwoNN intrinsic dimension estimator}\label{twonn_appendix}
The TwoNN method \citep{facco2017estimating} used in this work is a local intrinsic dimension ($I_d$) estimation technique. Its distinguishing feature is that, by focusing solely on the first two nearest neighbors, it minimizes the size of the $I_d$-dimensional hyperspheres over which the density is assumed to be constant. This method is based on the distribution functions of neighborhood distances, which depend on $I_d$.

Specifically, for each point $x$ in the dataset, it considers the first and second nearest-neighbor distances, denoted as $r_1(x)$ and $r_2(x)$, respectively. Assuming that the dataset is locally uniform within the range of second nearest neighbors, it has been shown by \citet{facco2017estimating} that the distribution of the ratio $\mu = r_2(x) / r_1(x)$ follows:

\begin{align}
f(\mu) = I_d \mu^{-I_d - 1}.    
\label{eq:fmu}
\end{align}

And using the corresponding cumulative distribution function, $P(\mu)$, one can write:

\begin{align}
 I_d = - \frac{\ln\left[ 1 -  P(\mu) \right]}{\ln\left( \mu \right)},
 \label{Id}
\end{align}

This relation allows for the estimation of $I_d$ by fitting the set $S = \{ (\ln(\mu), - \ln\left[ 1 - P^\textup{emp}(\mu) \right]   \}$
 with a straight line passing through the origin. Here, $P^\text{emp}(\mu)$ represents the empirical cumulative distribution, computed by sorting the values of $\mu$ in ascending order.

\subsection{Correlation through the intrinsic dimension: An exact example.}\label{example_id}
In order to provide an intuition of the relationship between the intrinsic dimension and the entropy of a data set, let us imagine data sets generated in this way: 1) Sample a multivariate Gaussian of dimension $d$ and identity covariance matrix. This can be done by simply generating $d$ series of Gaussian distributed numbers with standard deviation equal to 1. 2) Embed them in a higher dimensional space of dimension $D$ and apply a random angle rotation around a random vector of this dimension. 

The entropy of such a dataset could be approximated by the differential entropy of a multivariate Gaussian distribution
\begin{equation}
    H=\frac{d}{2}\log(2\pi e)+ \frac{1}{2}\log\det(Cov)=\frac{d}{2}\log(2\pi e)
\end{equation}
where the last equality derives from the use of an identity covariance matrix.

Another interesting property of such a dataset is that, since we have generated the dataset through rotation, the quantity $d$ can be recovered as the number of non-zero eigenvalues of the covariance matrix. Please note that this is equivalent to identifying the intrinsic dimension using PCA.

Now let us imagine that we generate three datasets in this way ($X_1$, $X_2$ and $X_3$) each of them with its own intrinsic dimension $d_1$, $d_2$, $d_3$ and embedding dimension $D_1$, $D_2$, $D_3$. While datasets $X_1$ and $X_2$ are completely independent, dataset $X_3$ is built using one of the series employed for building dataset $X_1$.

We will use PCA for estimating the $I_d$ and apply equation \ref{idcor_eq} to compute the correlation between datasets these datasets.
In the case of datasets $X_1$ and $X_2$, being fully independent, we will have $I_d(X_1)=d_1$, $I_d(X_2)=d_2$, and $I_d(X_1\oplus X_2)=d_1+d_2$. By applying PCA, we can recover our datasets as multivariate normal Gaussian, therefore obtaining: 
\begin{multline}
\varrho = \frac{H(X_1)+H(X_2)-H(X_1\oplus X_2)}{\max(H(X_1), H(X_2))} =
\frac{\frac{d_1}{2}\log(2\pi e)+\frac{d_2}{2}\log(2\pi e)-\frac{d_1+d_2}{2}\log(2\pi e)}{\max(\frac{d_1}{2}\log(2\pi e), \frac{d_2}{2}\log(2\pi e))} =\\ 
\frac{d_1+d_2-(d_1+d_2))}{\max(d_1, d_2)}=
I_{d}\text{Cor}(X_1,X_2) =0
\end{multline}
The situation is identical for datasets $X_2$ and $X_3$ but, for $X_1$ and $X_3$, the situation changes a bit. In this case, applying PCA to the combined dataset $X_1\oplus X_3$ will also provide a multivariate Gaussian, but with a dimension equal to $d_1+d_3-1$, therefore obtaining $\varrho=I_d\text{Cor}(X_1,X_3)=\frac{1}{\max(d_1, d_3)}$.
\subsection{Additional results on synthetic data}
\begin{figure}[h]
  \centering
  \includegraphics[width=\textwidth]{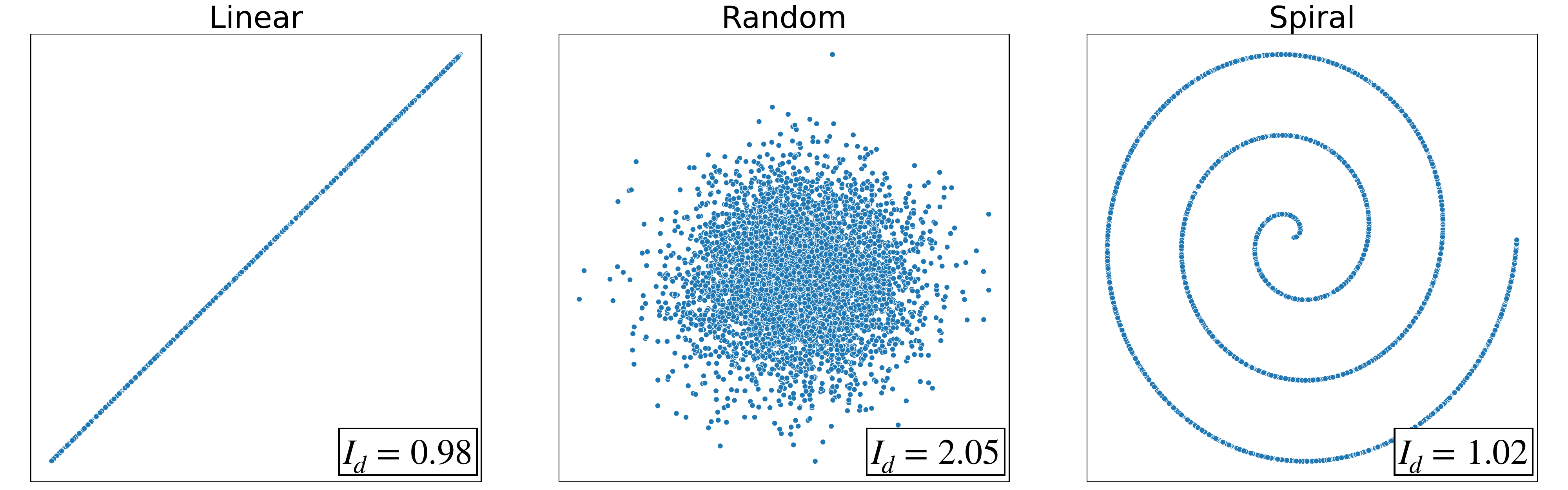}
  \caption{Synthetic datasets, each associated with its intrinsic dimension.}\label{toyexample}
\end{figure}

In Fig. \ref{toyexample}, we display the two-variable synthetic datasets we used in section \ref{2dexp} as a first validation for our method.

\subsubsection{Results using the MLE estimator}\label{mle_appendix}

\begin{figure}[h]
  \centering
  \includegraphics[width=0.8\textwidth]{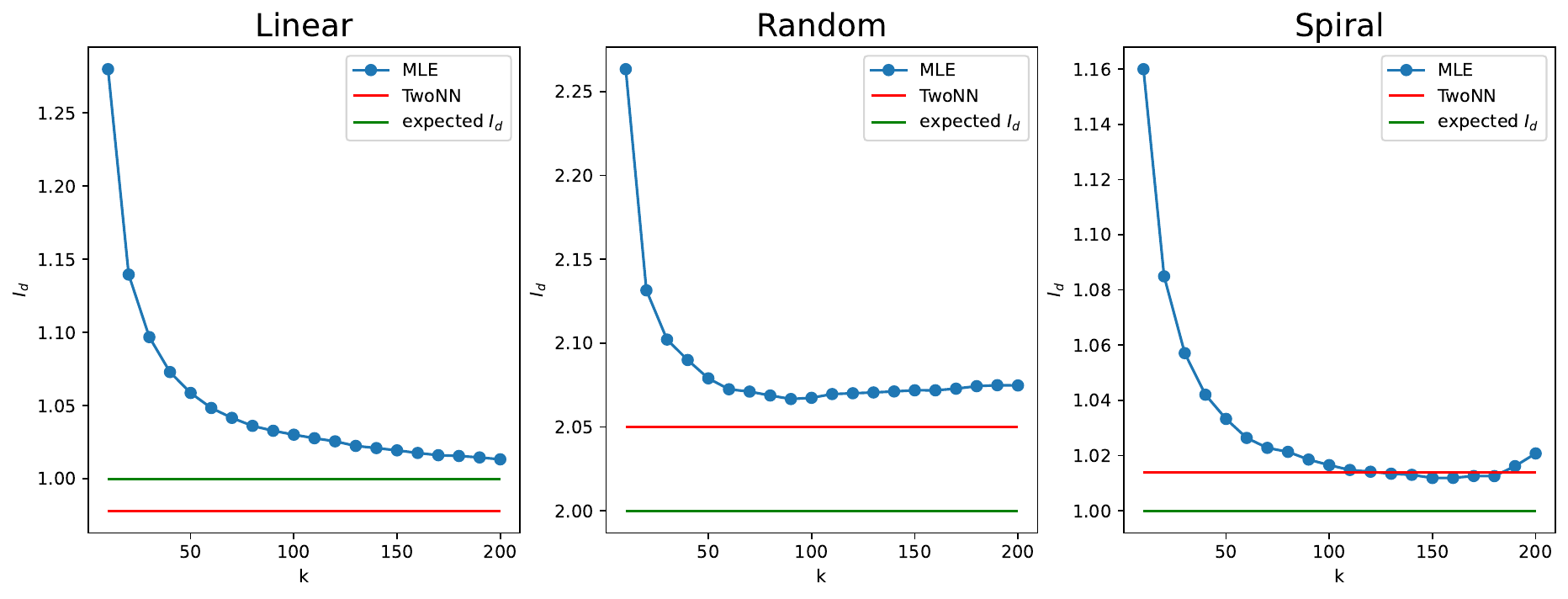}
  \caption{Intrinsic dimension estimation with the MLE estimator, varying the number of neearest neighbors $k$. Horizontal red and green lines report the estimate found by TwoNN and the expected $I_d$ of data, respectively.}\label{vark}
\end{figure}

$I_d$Cor is fundamentally powered by an intrinsic dimension estimator. In the main text, we employed TwoNN \citep{facco2017estimating}, but this is only one of the possible choices. To highlight this point, we provide here results on the synthetic experiments of section \ref{2dexp}, using the Maximum Likelihood Estimator (MLE) by \citet{levina2004maximum} instead of TwoNN. MLE relies fundamentally on a hyperparameter $k$, the number of nearest neighbors to consider for each data point. As we show in Fig. \ref{vark}, $k$ can significantly impact the estimated $I_d$ and, in general, there is no clear strategy to choose it. Based on these curves, we opt for $k=100$ in our experiments, to obtain a reasonable trade-off between efficiency (the lower $k$ the faster MLE is) and accuracy. We report in Table \ref{toy_results_appendix} the $I_d$Cor results with MLE in the same datasets of Table \ref{toy_results}.
\begin{table}[h]
\centering 
  \caption{Correlation in the two-variables datasets of Table \ref{toy_results}.
   ({$\mathbf{I_d \oplus}$}): intrinsic dimension of the concatenated (2D) dataset, computed with MLE; ($\mathbf{I_dCor}$): correlation coefficient ($\varrho$) returned by our method; ($\mathbf{p}$\textbf{-value}): significance of the correlation detected by our method ($100$ shuffles). Results are reported as $\text{mean}\pm\text{std}$ over 10 independent random samplings of data, with the exclusion of $p$-value, which is reported as a range.\\}
  \label{toy_results_appendix}
  \centering
  \begin{tabular}{cccc}
    \toprule
    \textbf{Data} &  $\mathbf{I_d \oplus}$ & $\mathbf{I_dCor}$ & $\mathbf{p}$\textbf{-value}\\
    \midrule
    Linear  & $1.03\pm0.00$ & $1.00\pm0.00$ & $0.01$  \\
    Random  & $2.07\pm0.00$ & $0.01\pm0.00$ & $[0.04, 0.76]$  \\
    Spiral  & $1.02\pm0.00$ & $1.06\pm0.00$ & $0.01$  \\
    \bottomrule
  \end{tabular}
\end{table}

In Table \ref{4d_results_appendix}, we report the results of $I_d$Cor equipped with MLE $I_d$ estimation on the same datasets employed in Table \ref{4d_results}. We observe that $I_d$Cor equipped with MLE produces slightly worse results with respect to TwoNN, as $I_d\oplus$ is estimated less accurately (expected values would be 8, 6 and 4 respectively for scenarios A, B and C).
\begin{table}[h]
\centering 
  \caption{Correlation between high dimensional synthetic datasets in the scenarios of Table \ref{4d_results}, when TwoNN is replaced by MLE in the computation of $I_d$Cor. Results are reported as $\text{mean}\pm\text{std}$ over 10 independent random samplings of data, with the exclusion of $p$-value, which is reported as a range.\\}
  \label{4d_results_appendix}
  \centering
  \begin{tabular}{cccc}
    \toprule
    \textbf{Correlation} &  $\mathbf{I_d \oplus}$& $\mathbf{I_d Cor}$ & $\mathbf{p}$\textbf{-value}\\
    \midrule
    (A) Absent & $7.52\pm0.01$ & $0.17\pm0.00$ & $[0.17, 0.93]$ \\
    (B) Partial & $6.68\pm0.01$ & $0.27\pm0.00$ & $0.01$ \\
    (C) Complete & $5.81\pm0.01$ & $0.38\pm0.01$ & $0.01$ \\
    \bottomrule
  \end{tabular}
\end{table}
\newpage
\subsubsection{Results on a different data distribution}\label{uniform_appendix}
In Table \ref{4d_results_uniform}, we provide the results for the same experiment of Table \ref{4d_results}, when data are sample from a uniform distribution in $[-\pi, \pi]$ instead of a Gaussian distribution. From this point, we switch back to the TwoNN estimator.
\begin{table}[h]
\centering 
  \caption{Correlation between high dimensional synthetic datasets in the scenarios of Table \ref{4d_results}, when the original dataset is sampled independently and uniformly in $[-\pi, \pi]$. Results are reported as $\text{mean}\pm\text{std}$ over 10 independent random samplings of data, with the exclusion of $p$-value, which is reported as a range.\\}
  \label{4d_results_uniform}
  \centering
  \begin{tabular}{cccccccc}
    \toprule
    \textbf{Correlation} & $\mathbf{CCA}$ & $\mathbf{dCor}$ &  $\mathbf{I_d \oplus}$& $\mathbf{I_d Cor}$ & $\mathbf{p}$\textbf{-value}\\
    \midrule
    (A) Absent & $0.02\pm0.01$ & $0.00\pm0.00$ & $7.21\pm0.12$ & $0.15\pm0.04$ & $[0.09, 0.96]$ \\
    (B) Partial & $0.17\pm0.05$ & $0.06\pm0.00$ & $5.74\pm0.11$ & $0.48\pm0.04$ & $0.01$ \\
    (C) Complete & $0.23\pm0.02$ & $0.21\pm0.00$ & $4.03\pm0.06$ & $0.92\pm0.02$ & $0.01$ \\
    \bottomrule
  \end{tabular}
\end{table}
\newpage
\subsubsection{Robustness to noise}\label{noise_appendix}
\begin{figure}[h]
  \centering
\includegraphics[width=0.8\textwidth]{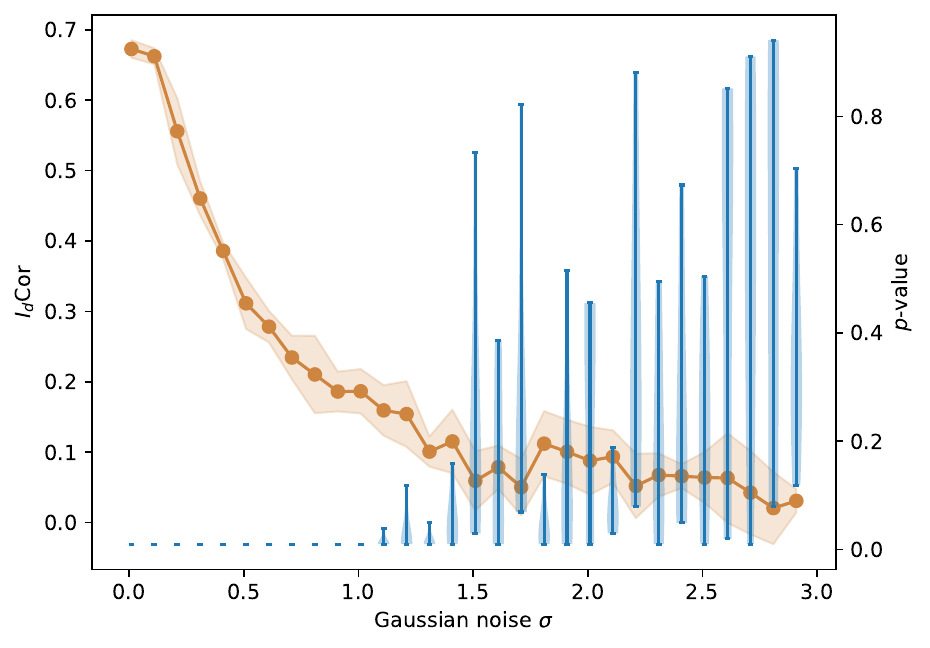}
  \caption{$I_d$Cor results (correlation coefficient and $p$-value) on completely correlated data from Table \ref{4d_results}, scenario \textbf{C}, perturbed with random Gaussian noise of variable magnitude.}\label{noisy_corr}
\end{figure}

To evaluate how $I_d$Cor behaves in a more complex and noisy setting, we consider Scenario \textbf{C (complete correlation)} from Table \ref{4d_results}, and perturb the second dataset $(x', y',w', z')$ with Gaussian noise with mean $0$ and increasing standard deviation, up to $3$ times larger than the scale of the dataset (all variables are obtained through sinusoidal functions, hence they are constrained in $[-1,1]$. We report in Fig. \ref{noisy_corr} the correlation results obtained with $I_d$Cor, averaging the correlation over $5$ independent random noise samplings for each standard deviation value. We also report the range of $p$-values for each noise magnitude, showing that correlation is detected with high confidence approximately until the point where noise and signal have similar amplitude.
\newpage
\subsection{Additional ImageNet correlation results}
\begin{figure}[h]
  \centering
  \includegraphics[width=\textwidth]{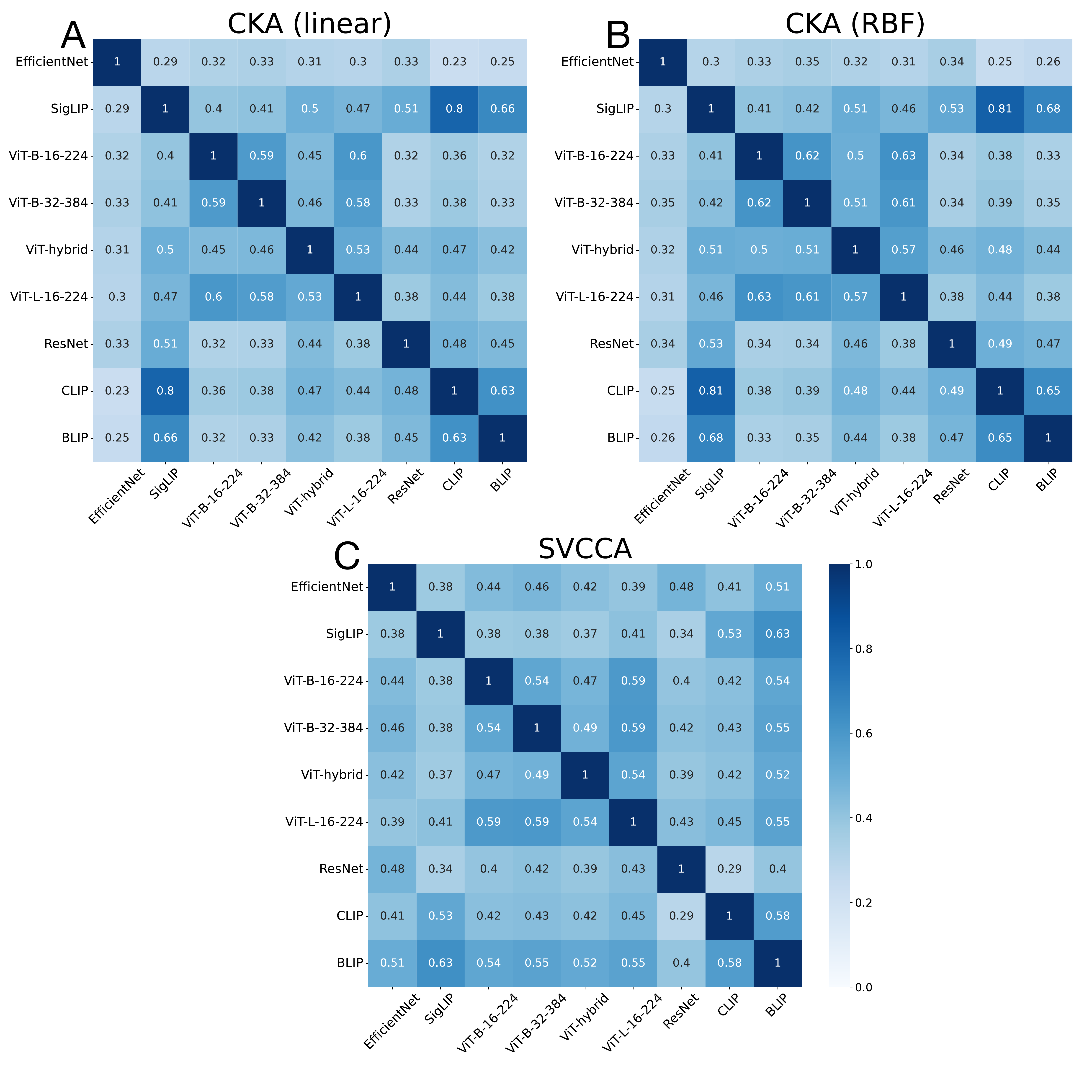}
  \caption{Additional baseline correlation results on ImageNet representations, obtained using: (A) Linear CKA; (B) RBF CKA; (C) SVCCA}\label{imagenetbaselines}
\end{figure}

We provide detailed results for latent correlation in ImageNet (Fig. \ref{imagenetbaselines}), using SVCCA, linear kernel CKA and RBF kernel CKA.

\newpage

\subsection{Additional correlation results on multimodal datasets}\label{additional}
\subsubsection{Additional baseline results on N24News}\label{n24appendix}
\begin{figure}[h]
  \centering
  \includegraphics[width=\textwidth]{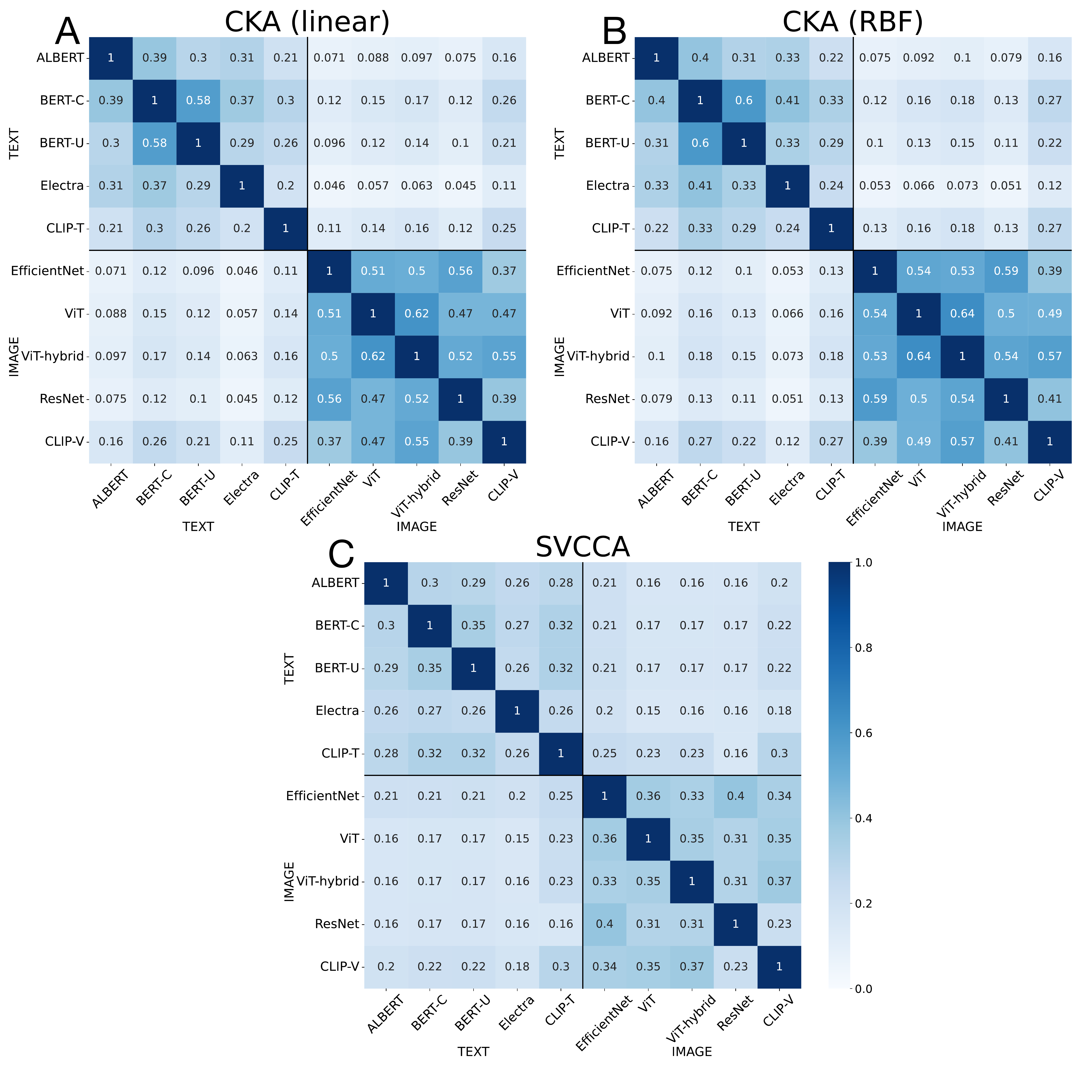}
  \caption{Additional baseline correlation results on N24News representations, obtained using: (A) Linear CKA; (B) RBF CKA; (C) SVCCA}\label{n24baselines}
\end{figure}

In Fig. \ref{n24baselines}, we provide additional results for SVCCA, linear kernel CKA and RBF kernel CKA on N24News representations. These methods perform similarly to Distance Correlation, and only capture high correlation between representations belonging to the same data modality.
\newpage
\subsubsection{Results on other multimodal datasets}\label{flickrcoco}
\begin{figure}[h]
  \centering
  \includegraphics[width=0.8\textwidth]{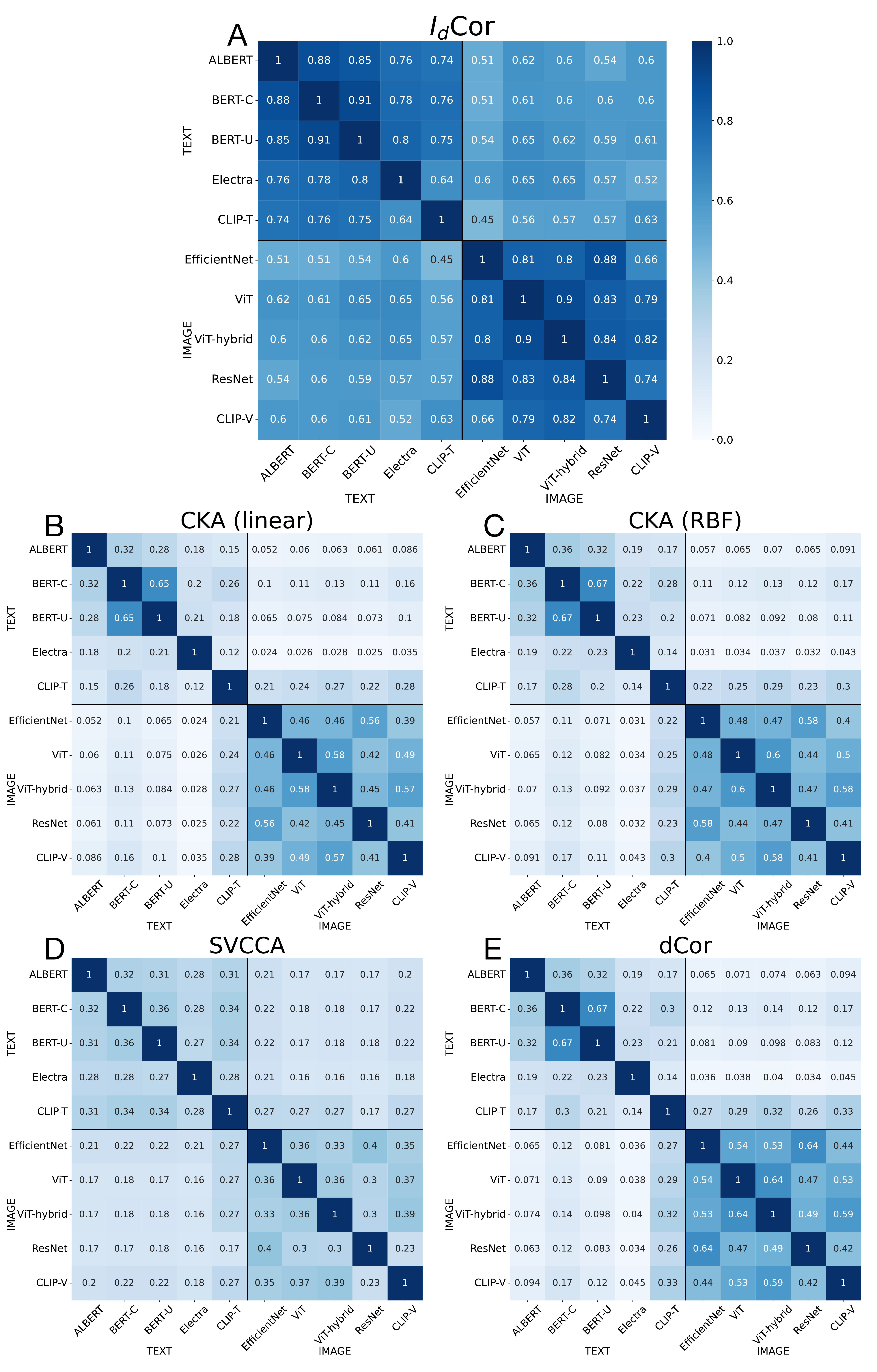}
  \caption{Correlation results on Flickr30k representations, obtained using: (A) our method $I_d$Cor; (B) linear kernel CKA; (C) RBF kernel CKA; (D) SVCCA; (E) Distance Correlation}\label{flickrcorrelation}
\end{figure}

\begin{figure}[h]
  \centering
  \includegraphics[width=0.8\textwidth]{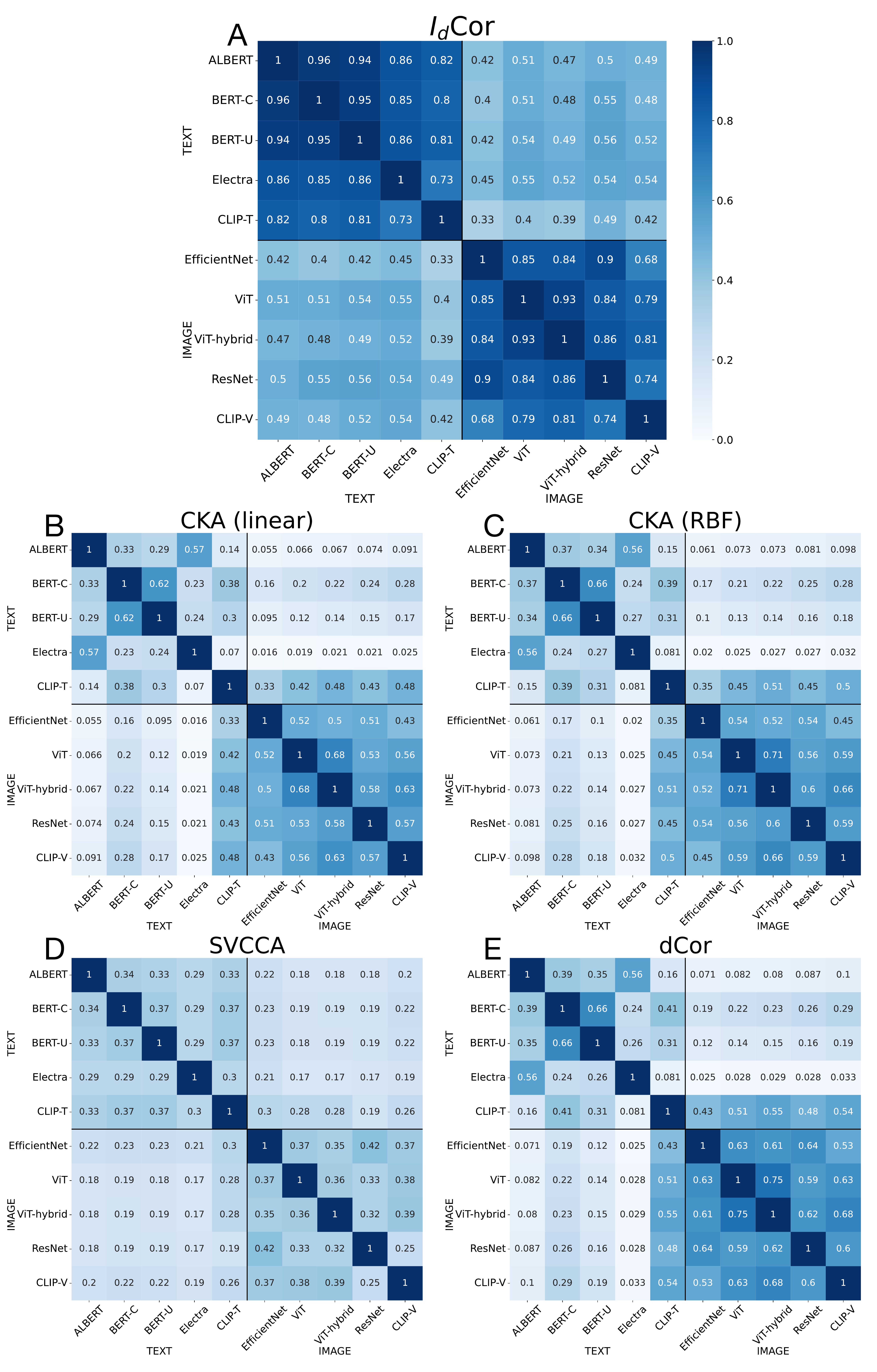}
  \caption{Correlation results on MS-COCO representations, obtained using: (A) our method $I_d$Cor; (B) linear kernel CKA; (C) RBF kernel CKA; (D) SVCCA; (E) Distance Correlation}\label{cococorrelation}
\end{figure}

In Fig. \ref{flickrcorrelation} and Fig. \ref{cococorrelation} we report the correlation results we obtain respectively on Flickr30k \citep{young2014image} and MS-COCO \citep{lin2014microsoft}, two multimodal datasets containing images and the corresponding captions. We employ the same models used for N24News (section \ref{n24news}). As for the previous dataset, we observe that $I_d$Cor outperforms previous baselines, which significantly struggle to find similarities between embeddings of different modalities.
\newpage
\newpage
\subsubsection{Partial multimodal correlation}\label{5captionsappendix}
\begin{figure}[h]
  \centering
  \includegraphics[width=0.8\textwidth]{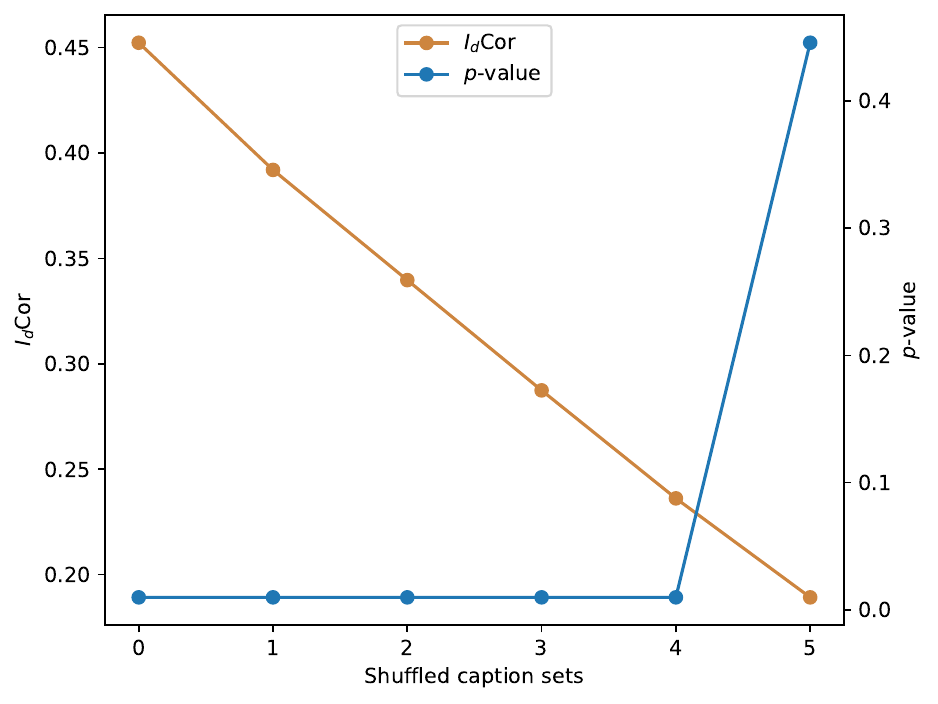}
  \caption{Correlation between visual and textual representations in Flickr30k, considering all 5 captions per image and shuffling an increasing number of textual encoding sets. Representations are computed using CLIP-ViT (images) and BERT (text). }\label{captions_shuffle}
\end{figure}

In a multimodal setting, $I_d$Cor adapts to the strength of the correlation between textual and visual representation. In this experiment, we consider again the Flickr30k dataset, which provides 5 human annotated captions per image. Instead of computing the correlation between each image representation and a single corresponding caption set, we consider here an enriched textual representation, containing the concatenated encoding for all 5 captions. Then, we progressively perturb the correlation by shuffling an increasing number of textual representations. As we report in Fig. \ref{captions_shuffle}, $I_d$Cor decreases almost linearly with the number of perturbed textual representations. Instead, the $p$-value stays at the minimum ($0.01$) until the last step, where all caption embeddings are shuffled and, consequently, correlation is totally lost.
\newpage

\subsection{Correlation between different datasets}\label{sanityappendix}
\begin{figure}[h]
  \centering
  \includegraphics[width=\textwidth]{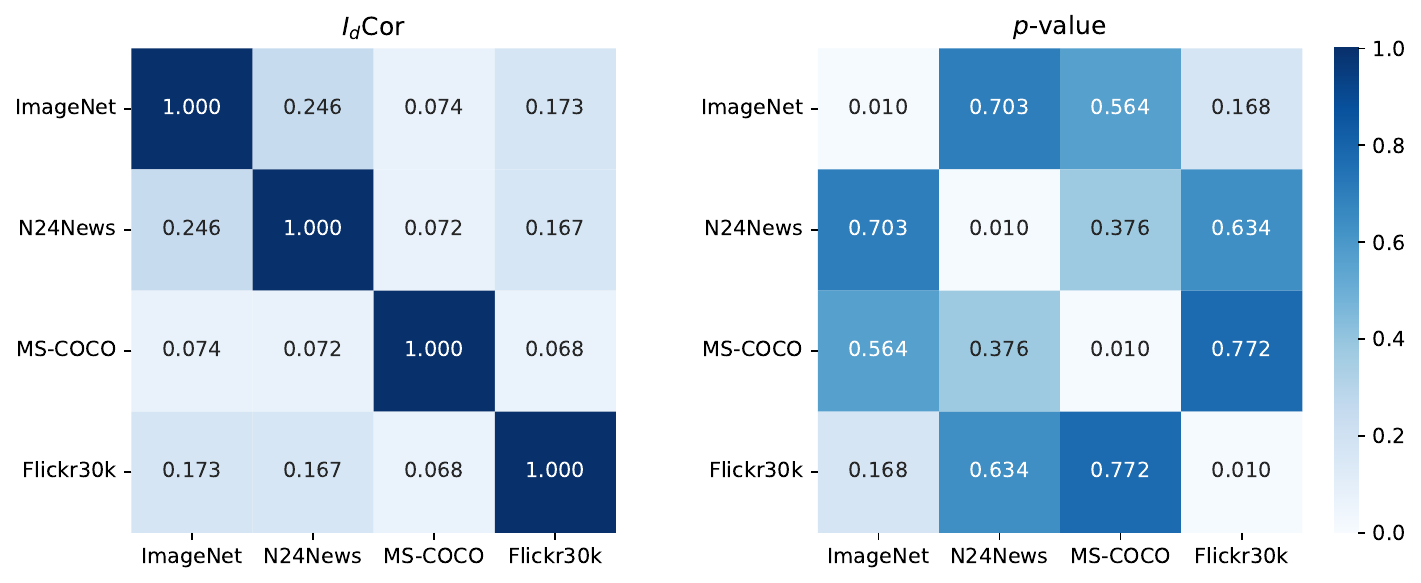}
  \caption{Correlation results between representations produced by the CLIP vision encoder on different image datasets, reported in terms of $I_d$Cor coefficient (left) and $p$-value (right).}\label{sanity_check}
\end{figure}

In this section, we provide an assessment of $I_d$Cor on unimodal image representations of different input datasets, computed using CLIP-ViT-B. As we report in Fig. \ref{sanity_check}, $I_d$Cor scores significantly fall off-diagonal (on different input data), and the corresponding $p$-values increase, indicating lack of correlation.

\subsection{Multimodal RTD scores}\label{rtd}
\begin{figure}[h]
  \centering
  \includegraphics[width=0.6\textwidth]{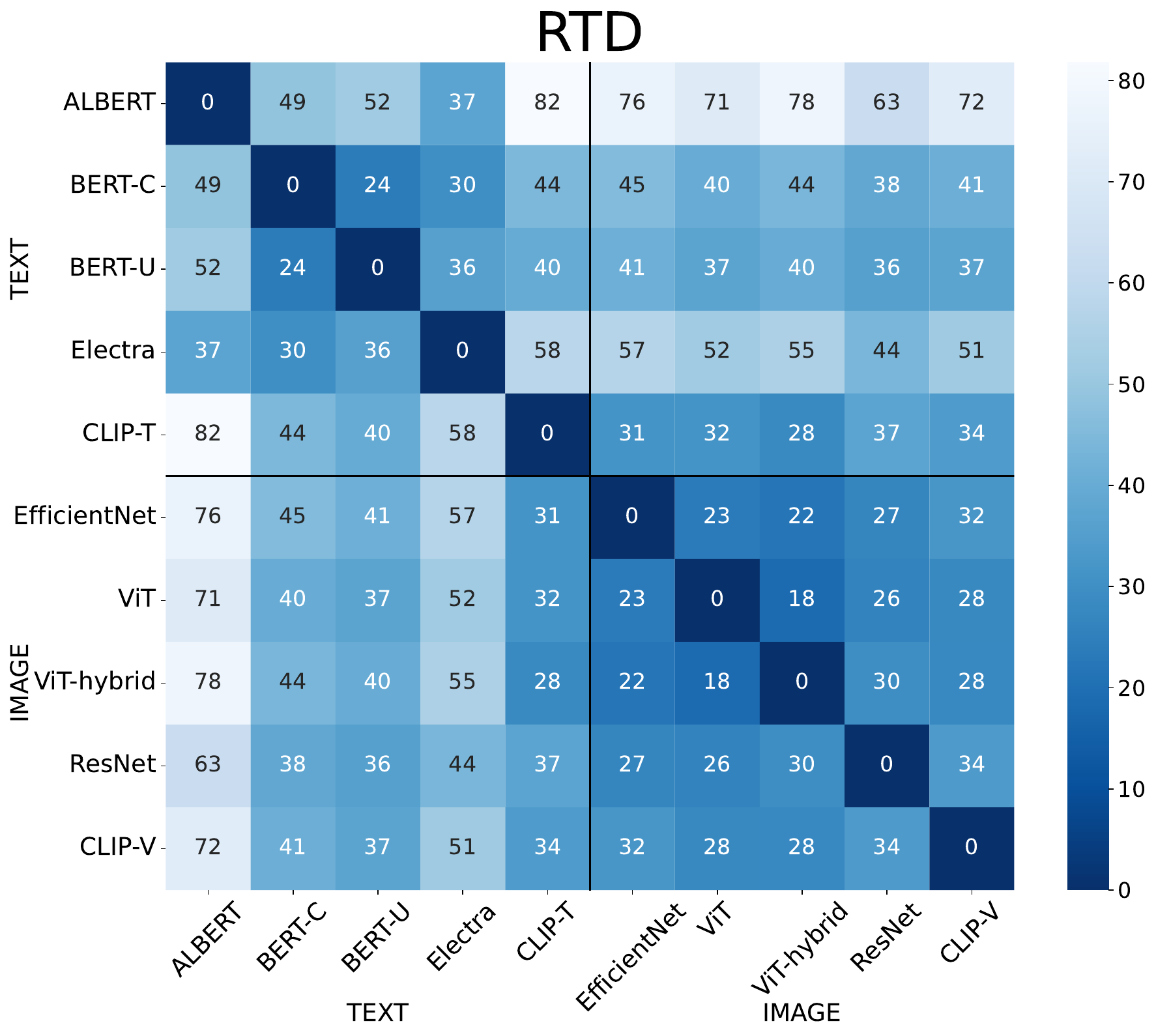}
  \caption{RTD \citep{barannikov2022representation} scores on N24News representations. Colormap is inverted w.r.t. previous figures as RTD produces a divergence score, and not a similarity score.}\label{results_rtd}
\end{figure}

In this section, we report the divergence scores computed using Representation Topology Divergence (RTD) \citep{barannikov2022representation} on N24News encodings. As reported in Fig. \ref{results_rtd}, RTD captures stronger similarities within the visual domain (off-diagonal mean divergence is $26.8$), while it finds weaker similarities in the textual and cross-modal settings, with means slightly above $45$. 

\subsection{Model details}\label{models}
All models we employ are taken pre-trained from the HuggingFace \texttt{transformers} \cite{wolf2020transformers} library. We report in Table \ref{modelguide} the full list of pre-trained models we employed in this work, associated with the name of the corresponding checkpoint in the library.
\begin{table}[h]

\centering 
\
  \caption{Reference guide for pre-trained model checkpoints in HuggingFace \texttt{transformers} \cite{wolf2020transformers} library.}\label{modelguide}
  \centering
  \begin{tabular}{ll}
    \toprule
    Name in the paper & pre-trained checkpoint name\\
    \midrule
    ALBERT & \texttt{albert/albert-base-v2}\\
    BERT-C & \texttt{google-bert/bert-base-cased}\\
    BERT-U & \texttt{google-bert/bert-base-uncased}\\
    Electra & \texttt{google/electra-base-discriminator}\\
    CLIP (-T/-V) & \texttt{openai/clip-vit-base-patch16}\\
    BLIP & \texttt{Salesforce/blip-itm-base-flickr}\\
    EfficientNet & \texttt{google/efficientnet-b0}\\
    SigLIP & \texttt{google/siglip-base-patch16-224}\\
    ViT-B-16-224 (ViT) & \texttt{google/vit-base-patch16-224}\\
    ViT-B-32-384 & \texttt{google/vit-base-patch32-384}\\
    ViT-hybrid & \texttt{google/vit-hybrid-base-bit-384}\\
    ViT-L-16-224 & \texttt{google/vit-large-patch16-224}\\
    ResNet & \texttt{microsoft/resnet-18}\\
    \bottomrule
  \end{tabular}

\end{table}
\newpage
\subsection{Local $I_d$Cor}\label{ian_appendix}
In this section, we provide a proof-of-concept illustration of a local version of $I_d$Cor, that employs IAN \citep{dyballa2023ian}, a local $I_d$ estimator.
\begin{figure}[h]
  \centering
  \includegraphics[width=0.8\textwidth]{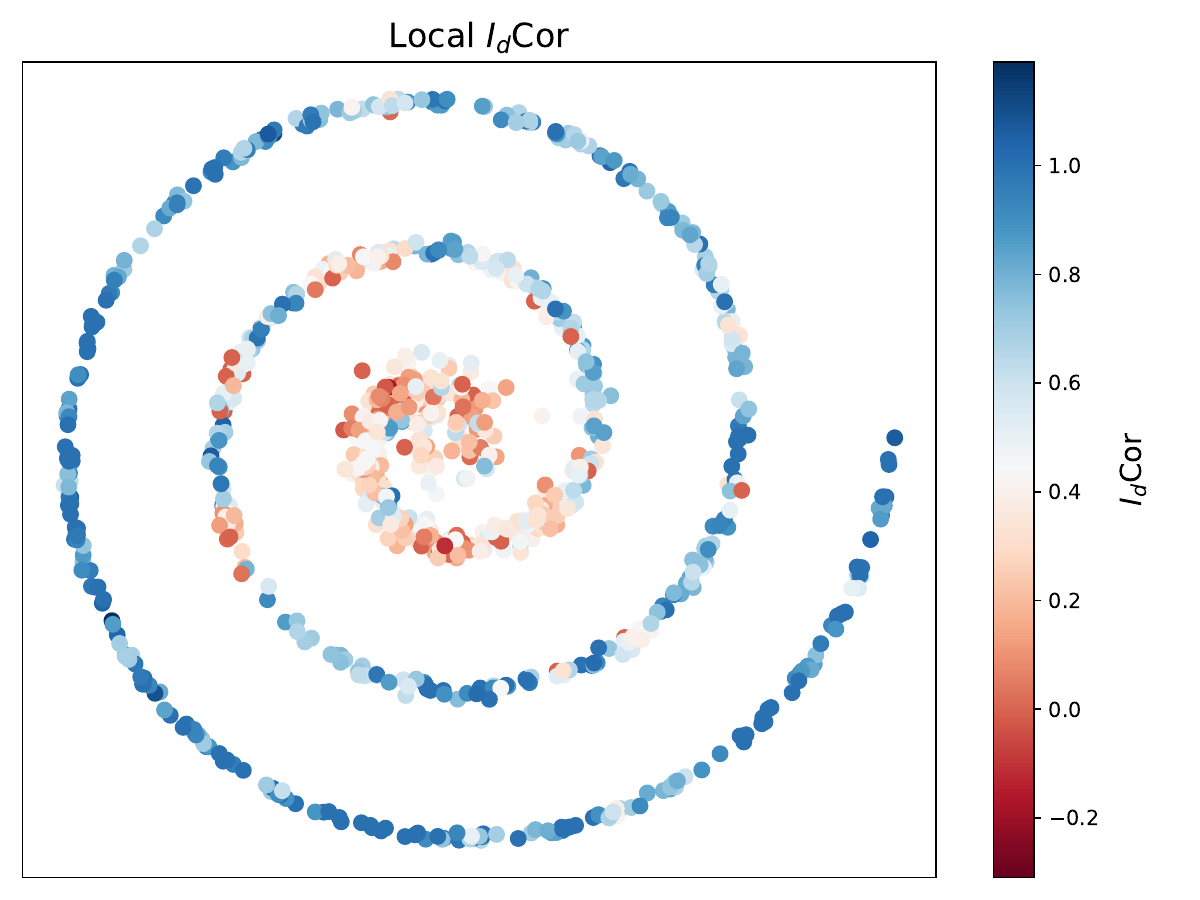}
  \caption{Local $I_d$Cor with IAN estimator. It can be seen that in the center of the spiral, where the data is nearly uniformly distributed, the correlation is low, while in the outer region, where correlations force the data to be in the arms of the spiral, the $I_d$Cor index is 1.}\label{local_idcor}
\end{figure}
\end{document}

%% file: math_commands.tex
\usepackage{amsmath,amsfonts,bm}

\def\eqref#1{equation~\ref{#1}}

\def\1{\bm{1}}

\DeclareMathAlphabet{\mathsfit}{\encodingdefault}{\sfdefault}{m}{sl}
\SetMathAlphabet{\mathsfit}{bold}{\encodingdefault}{\sfdefault}{bx}{n}













%% file: cameraready.bbl
\begin{thebibliography}{57}
\providecommand{\natexlab}[1]{#1}
\providecommand{\url}[1]{\texttt{#1}}
\expandafter\ifx\csname urlstyle\endcsname\relax
  \providecommand{\doi}[1]{doi: #1}\else
  \providecommand{\doi}{doi: \begingroup \urlstyle{rm}\Url}\fi

\bibitem[Allegra et~al.(2020)Allegra, Facco, Denti, Laio, and Mira]{allegra2020data}
Michele Allegra, Elena Facco, Francesco Denti, Alessandro Laio, and Antonietta Mira.
\newblock Data segmentation based on the local intrinsic dimension.
\newblock \emph{Scientific reports}, 10\penalty0 (1):\penalty0 16449, 2020.

\bibitem[Ansuini et~al.(2019)Ansuini, Laio, Macke, and Zoccolan]{ansuini2019intrinsic}
Alessio Ansuini, Alessandro Laio, Jakob~H Macke, and Davide Zoccolan.
\newblock Intrinsic dimension of data representations in deep neural networks.
\newblock \emph{Advances in Neural Information Processing Systems}, 32, 2019.

\bibitem[Bailey et~al.(2022)Bailey, Houle, and Ma]{bailey2022local}
James Bailey, Michael~E Houle, and Xingjun Ma.
\newblock Local intrinsic dimensionality, entropy and statistical divergences.
\newblock \emph{Entropy}, 24\penalty0 (9):\penalty0 1220, 2022.

\bibitem[Barannikov et~al.(2022)Barannikov, Trofimov, Balabin, and Burnaev]{barannikov2022representation}
Serguei Barannikov, Ilya Trofimov, Nikita Balabin, and Evgeny Burnaev.
\newblock Representation topology divergence: A method for comparing neural network representations.
\newblock In \emph{International Conference on Machine Learning}, pp.\  1607--1626. PMLR, 2022.

\bibitem[Brown et~al.(2022)Brown, Juravsky, Caterini, and Loaiza-Ganem]{brown2022relating}
Bradley~CA Brown, Jordan Juravsky, Anthony~L Caterini, and Gabriel Loaiza-Ganem.
\newblock Relating regularization and generalization through the intrinsic dimension of activations.
\newblock In \emph{OPT 2022: Optimization for Machine Learning (NeurIPS 2022 Workshop)}, 2022.

\bibitem[Campadelli et~al.(2015)Campadelli, Casiraghi, Ceruti, and Rozza]{campadelli2015intrinsic}
Paola Campadelli, Elena Casiraghi, Claudio Ceruti, and Alessandro Rozza.
\newblock Intrinsic dimension estimation: Relevant techniques and a benchmark framework.
\newblock \emph{Mathematical Problems in Engineering}, 2015:\penalty0 1--21, 2015.

\bibitem[Ceruti et~al.(2014)Ceruti, Bassis, Rozza, Lombardi, Casiraghi, and Campadelli]{ceruti2014danco}
Claudio Ceruti, Simone Bassis, Alessandro Rozza, Gabriele Lombardi, Elena Casiraghi, and Paola Campadelli.
\newblock Danco: An intrinsic dimensionality estimator exploiting angle and norm concentration.
\newblock \emph{Pattern recognition}, 47\penalty0 (8):\penalty0 2569--2581, 2014.

\bibitem[Cheng et~al.(2023)Cheng, Kervadec, and Baroni]{cheng2023bridging}
Emily Cheng, Corentin Kervadec, and Marco Baroni.
\newblock Bridging information-theoretic and geometric compression in language models.
\newblock In \emph{Proceedings of the 2023 Conference on Empirical Methods in Natural Language Processing}, pp.\  12397--12420, 2023.

\bibitem[Ciernik et~al.(2024)Ciernik, Linhardt, Morik, Dippel, Kornblith, and Muttenthaler]{ciernik2024training}
Laure Ciernik, Lorenz Linhardt, Marco Morik, Jonas Dippel, Simon Kornblith, and Lukas Muttenthaler.
\newblock Training objective drives the consistency of representational similarity across datasets.
\newblock \emph{arXiv preprint arXiv:2411.05561}, 2024.

\bibitem[Clark et~al.(2020)Clark, Luong, Le, and Manning]{clark2020electra}
Kevin Clark, Minh-Thang Luong, Quoc~V. Le, and Christopher~D. Manning.
\newblock {ELECTRA}: Pre-training text encoders as discriminators rather than generators.
\newblock In \emph{ICLR}, 2020.
\newblock URL \url{https://openreview.net/pdf?id=r1xMH1BtvB}.

\bibitem[Davari et~al.(2023)Davari, Horoi, Natik, Lajoie, Wolf, and Belilovsky]{davari2023reliability}
MohammadReza Davari, Stefan Horoi, Amine Natik, Guillaume Lajoie, Guy Wolf, and Eugene Belilovsky.
\newblock Reliability of {CKA} as a similarity measure in deep learning.
\newblock In \emph{The Eleventh International Conference on Learning Representations}, 2023.
\newblock URL \url{https://openreview.net/forum?id=8HRvyxc606}.

\bibitem[Davison \& Hinkley(1997)Davison and Hinkley]{davison1997bootstrap}
Anthony~Christopher Davison and David~Victor Hinkley.
\newblock \emph{Bootstrap methods and their application}.
\newblock Number~1. Cambridge university press, 1997.

\bibitem[Devlin et~al.(2019)Devlin, Chang, Lee, and Toutanova]{devlin2019bert}
Jacob Devlin, Ming-Wei Chang, Kenton Lee, and Kristina Toutanova.
\newblock Bert: Pre-training of deep bidirectional transformers for language understanding.
\newblock In \emph{Proceedings of the 2019 Conference of the North American Chapter of the Association for Computational Linguistics: Human Language Technologies, Volume 1 (Long and Short Papers)}, pp.\  4171--4186, 2019.

\bibitem[Dosovitskiy et~al.(2020)Dosovitskiy, Beyer, Kolesnikov, Weissenborn, Zhai, Unterthiner, Dehghani, Minderer, Heigold, Gelly, et~al.]{dosovitskiy2020image}
Alexey Dosovitskiy, Lucas Beyer, Alexander Kolesnikov, Dirk Weissenborn, Xiaohua Zhai, Thomas Unterthiner, Mostafa Dehghani, Matthias Minderer, Georg Heigold, Sylvain Gelly, et~al.
\newblock An image is worth 16x16 words: Transformers for image recognition at scale.
\newblock In \emph{International Conference on Learning Representations}, 2020.

\bibitem[Dyballa \& Zucker(2023)Dyballa and Zucker]{dyballa2023ian}
Luciano Dyballa and Steven~W Zucker.
\newblock Ian: Iterated adaptive neighborhoods for manifold learning and dimensionality estimation.
\newblock \emph{Neural Computation}, 35\penalty0 (3):\penalty0 453--524, 2023.

\bibitem[Facco et~al.(2017)Facco, d’Errico, Rodriguez, and Laio]{facco2017estimating}
Elena Facco, Maria d’Errico, Alex Rodriguez, and Alessandro Laio.
\newblock Estimating the intrinsic dimension of datasets by a minimal neighborhood information.
\newblock \emph{Scientific reports}, 7\penalty0 (1):\penalty0 12140, 2017.

\bibitem[Ghosh \& Motani(2023)Ghosh and Motani]{ghosh2023local}
Rohan Ghosh and Mehul Motani.
\newblock Local intrinsic dimensional entropy.
\newblock In \emph{Proceedings of the AAAI Conference on Artificial Intelligence}, volume~37, pp.\  7714--7721, 2023.

\bibitem[Hamilton et~al.(2016)Hamilton, Leskovec, and Jurafsky]{hamilton2016diachronic}
William~L Hamilton, Jure Leskovec, and Dan Jurafsky.
\newblock Diachronic word embeddings reveal statistical laws of semantic change.
\newblock In \emph{Proceedings of the 54th Annual Meeting of the Association for Computational Linguistics (Volume 1: Long Papers)}, pp.\  1489--1501, 2016.

\bibitem[He et~al.(2016)He, Zhang, Ren, and Sun]{he2016deep}
Kaiming He, Xiangyu Zhang, Shaoqing Ren, and Jian Sun.
\newblock Deep residual learning for image recognition.
\newblock In \emph{Proceedings of the IEEE conference on computer vision and pattern recognition}, pp.\  770--778, 2016.

\bibitem[Horibe(1985)]{horibe1985entropy}
Yasuichi Horibe.
\newblock Entropy and correlation.
\newblock \emph{IEEE Transactions on Systems, Man, and Cybernetics}, \penalty0 (5):\penalty0 641--642, 1985.

\bibitem[Hotelling(1936)]{hotelling1936relations}
Harold Hotelling.
\newblock Relations between two sets of variates.
\newblock \emph{Biometrika}, 28\penalty0 (3/4):\penalty0 321--377, 1936.

\bibitem[Kaufman \& Azencot(2023)Kaufman and Azencot]{kaufman2023data}
Ilya Kaufman and Omri Azencot.
\newblock Data representations’ study of latent image manifolds.
\newblock In \emph{International Conference on Machine Learning}, pp.\  15928--15945. PMLR, 2023.

\bibitem[Klabunde et~al.(2023)Klabunde, Schumacher, Strohmaier, and Lemmerich]{klabunde2023similarity}
Max Klabunde, Tobias Schumacher, Markus Strohmaier, and Florian Lemmerich.
\newblock Similarity of neural network models: A survey of functional and representational measures.
\newblock \emph{arXiv preprint arXiv:2305.06329}, 2023.

\bibitem[Kornblith et~al.(2019)Kornblith, Norouzi, Lee, and Hinton]{kornblith2019similarity}
Simon Kornblith, Mohammad Norouzi, Honglak Lee, and Geoffrey Hinton.
\newblock Similarity of neural network representations revisited.
\newblock In \emph{International conference on machine learning}, pp.\  3519--3529. PMLR, 2019.

\bibitem[Kvinge et~al.(2023)Kvinge, Brown, and Godfrey]{kvinge2023exploring}
Henry Kvinge, Davis Brown, and Charles Godfrey.
\newblock Exploring the representation manifolds of stable diffusion through the lens of intrinsic dimension.
\newblock \emph{arXiv preprint arXiv:2302.09301}, 2023.

\bibitem[L\"ahner \& Moeller(2024)L\"ahner and Moeller]{lahner2023direct}
Zorah L\"ahner and Michael Moeller.
\newblock On the direct alignment of latent spaces.
\newblock In \emph{Proceedings of UniReps: the First Workshop on Unifying Representations in Neural Models}, volume 243 of \emph{Proceedings of Machine Learning Research}, pp.\  158--169. PMLR, 2024.
\newblock URL \url{https://proceedings.mlr.press/v243/lahner24a.html}.

\bibitem[Lan et~al.(2019)Lan, Chen, Goodman, Gimpel, Sharma, and Soricut]{lan2019albert}
Zhenzhong Lan, Mingda Chen, Sebastian Goodman, Kevin Gimpel, Piyush Sharma, and Radu Soricut.
\newblock Albert: A lite bert for self-supervised learning of language representations.
\newblock In \emph{International Conference on Learning Representations}, 2019.

\bibitem[LeCun et~al.(1998)LeCun, Bottou, Bengio, and Haffner]{lecun1998gradient}
Yann LeCun, L{\'e}on Bottou, Yoshua Bengio, and Patrick Haffner.
\newblock Gradient-based learning applied to document recognition.
\newblock \emph{Proceedings of the IEEE}, 86\penalty0 (11):\penalty0 2278--2324, 1998.

\bibitem[Lenc \& Vedaldi(2015)Lenc and Vedaldi]{lenc2015understanding}
Karel Lenc and Andrea Vedaldi.
\newblock Understanding image representations by measuring their equivariance and equivalence.
\newblock In \emph{Proceedings of the IEEE conference on computer vision and pattern recognition}, pp.\  991--999, 2015.

\bibitem[Levina \& Bickel(2004)Levina and Bickel]{levina2004maximum}
Elizaveta Levina and Peter Bickel.
\newblock Maximum likelihood estimation of intrinsic dimension.
\newblock \emph{Advances in neural information processing systems}, 17, 2004.

\bibitem[Li et~al.(2022)Li, Li, Xiong, and Hoi]{li2022blip}
Junnan Li, Dongxu Li, Caiming Xiong, and Steven Hoi.
\newblock Blip: Bootstrapping language-image pre-training for unified vision-language understanding and generation.
\newblock In \emph{International conference on machine learning}, pp.\  12888--12900. PMLR, 2022.

\bibitem[Lin et~al.(2014)Lin, Maire, Belongie, Hays, Perona, Ramanan, Doll{\'a}r, and Zitnick]{lin2014microsoft}
Tsung-Yi Lin, Michael Maire, Serge Belongie, James Hays, Pietro Perona, Deva Ramanan, Piotr Doll{\'a}r, and C~Lawrence Zitnick.
\newblock Microsoft coco: Common objects in context.
\newblock In \emph{Computer Vision--ECCV 2014: 13th European Conference, Zurich, Switzerland, September 6-12, 2014, Proceedings, Part V 13}, pp.\  740--755. Springer, 2014.

\bibitem[Maiorca(2024)]{maiorca2024latentis}
Valentino Maiorca.
\newblock Latentis: Your gateway to latent space communication.
\newblock \url{https://github.com/Flegyas/latentis}, 2024.

\bibitem[Maiorca et~al.(2023)Maiorca, Moschella, Norelli, Fumero, Locatello, and Rodol{\`a}]{maiorca2023translation}
Valentino Maiorca, Luca Moschella, Antonio Norelli, Marco Fumero, Francesco Locatello, and Emanuele Rodol{\`a}.
\newblock Latent space translation via semantic alignment.
\newblock In \emph{Thirty-seventh Conference on Neural Information Processing Systems}, 2023.

\bibitem[Merullo et~al.(2023)Merullo, Castricato, Eickhoff, and Pavlick]{merullo2023linearly}
Jack Merullo, Louis Castricato, Carsten Eickhoff, and Ellie Pavlick.
\newblock Linearly mapping from image to text space.
\newblock In \emph{The Eleventh International Conference on Learning Representations}, 2023.

\bibitem[Miranda(2021)]{miranda2021ultimate_anatome}
Brando Miranda.
\newblock Ultimate anatome, the ultimate pytorch library to analyze internal representation of neural networks.
\newblock \url{https://github.com/brando90/ultimate-anatome}, 2021.

\bibitem[Moayeri et~al.(2023)Moayeri, Rezaei, Sanjabi, and Feizi]{moayeri2023text}
Mazda Moayeri, Keivan Rezaei, Maziar Sanjabi, and Soheil Feizi.
\newblock Text-to-concept (and back) via cross-model alignment.
\newblock In \emph{International Conference on Machine Learning}, pp.\  25037--25060. PMLR, 2023.

\bibitem[Morcos et~al.(2018)Morcos, Raghu, and Bengio]{morcos2018insights}
Ari Morcos, Maithra Raghu, and Samy Bengio.
\newblock Insights on representational similarity in neural networks with canonical correlation.
\newblock \emph{Advances in neural information processing systems}, 31, 2018.

\bibitem[Moschella et~al.(2023)Moschella, Maiorca, Fumero, Norelli, Locatello, and Rodol{\`a}]{moschella2023relative}
Luca Moschella, Valentino Maiorca, Marco Fumero, Antonio Norelli, Francesco Locatello, and Emanuele Rodol{\`a}.
\newblock Relative representations enable zero-shot latent space communication.
\newblock In \emph{The Eleventh International Conference on Learning Representations}, 2023.
\newblock URL \url{https://openreview.net/forum?id=SrC-nwieGJ}.

\bibitem[Nguyen et~al.(2021)Nguyen, Raghu, and Kornblith]{nguyen2021do}
Thao Nguyen, Maithra Raghu, and Simon Kornblith.
\newblock Do wide and deep networks learn the same things? uncovering how neural network representations vary with width and depth.
\newblock In \emph{International Conference on Learning Representations}, 2021.
\newblock URL \url{https://openreview.net/forum?id=KJNcAkY8tY4}.

\bibitem[Norelli et~al.(2023)Norelli, Fumero, Maiorca, Moschella, Rodola, and Locatello]{norelli2023asif}
Antonio Norelli, Marco Fumero, Valentino Maiorca, Luca Moschella, Emanuele Rodola, and Francesco Locatello.
\newblock Asif: Coupled data turns unimodal models to multimodal without training.
\newblock \emph{Advances in Neural Information Processing Systems}, 37, 2023.

\bibitem[Radford et~al.(2021)Radford, Kim, Hallacy, Ramesh, Goh, Agarwal, Sastry, Askell, Mishkin, Clark, et~al.]{radford2021learning}
Alec Radford, Jong~Wook Kim, Chris Hallacy, Aditya Ramesh, Gabriel Goh, Sandhini Agarwal, Girish Sastry, Amanda Askell, Pamela Mishkin, Jack Clark, et~al.
\newblock Learning transferable visual models from natural language supervision.
\newblock In \emph{International conference on machine learning}, pp.\  8748--8763. PMLR, 2021.

\bibitem[Raghu et~al.(2017)Raghu, Gilmer, Yosinski, and Sohl-Dickstein]{raghu2017svcca}
Maithra Raghu, Justin Gilmer, Jason Yosinski, and Jascha Sohl-Dickstein.
\newblock Svcca: Singular vector canonical correlation analysis for deep learning dynamics and interpretability.
\newblock \emph{Advances in neural information processing systems}, 30, 2017.

\bibitem[Raghu et~al.(2021)Raghu, Unterthiner, Kornblith, Zhang, and Dosovitskiy]{raghu2021vision}
Maithra Raghu, Thomas Unterthiner, Simon Kornblith, Chiyuan Zhang, and Alexey Dosovitskiy.
\newblock Do vision transformers see like convolutional neural networks?
\newblock \emph{Advances in neural information processing systems}, 34:\penalty0 12116--12128, 2021.

\bibitem[Romano et~al.(2016)Romano, Chelly, Nguyen, Bailey, and Houle]{romano2016measuring}
Simone Romano, Oussama Chelly, Vinh Nguyen, James Bailey, and Michael~E Houle.
\newblock Measuring dependency via intrinsic dimensionality.
\newblock In \emph{2016 23rd international conference on pattern recognition (ICPR)}, pp.\  1207--1212. IEEE, 2016.

\bibitem[Russakovsky et~al.(2015)Russakovsky, Deng, Su, Krause, Satheesh, Ma, Huang, Karpathy, Khosla, Bernstein, et~al.]{russakovsky2015imagenet}
Olga Russakovsky, Jia Deng, Hao Su, Jonathan Krause, Sanjeev Satheesh, Sean Ma, Zhiheng Huang, Andrej Karpathy, Aditya Khosla, Michael Bernstein, et~al.
\newblock Imagenet large scale visual recognition challenge.
\newblock \emph{International journal of computer vision}, 115:\penalty0 211--252, 2015.

\bibitem[Sz{\'e}kely et~al.(2007)Sz{\'e}kely, Rizzo, and Bakirov]{szekely2007measuring}
G{\'a}bor~J Sz{\'e}kely, Maria~L Rizzo, and Nail~K Bakirov.
\newblock Measuring and testing dependence by correlation of distances.
\newblock \emph{The Annals of Statistics}, pp.\  2769--2794, 2007.

\bibitem[Tan \& Le(2019)Tan and Le]{tan2019efficientnet}
Mingxing Tan and Quoc Le.
\newblock Efficientnet: Rethinking model scaling for convolutional neural networks.
\newblock In \emph{International conference on machine learning}, pp.\  6105--6114. PMLR, 2019.

\bibitem[Valeriani et~al.(2023)Valeriani, Doimo, Cuturello, Laio, Ansuini, and Cazzaniga]{valeriani2023geometry}
Lucrezia Valeriani, Diego Doimo, Francesca Cuturello, Alessandro Laio, Alessio Ansuini, and Alberto Cazzaniga.
\newblock The geometry of hidden representations of large transformer models.
\newblock \emph{Advances in Neural Information Processing Systems}, 37, 2023.

\bibitem[Vinh et~al.(2009)Vinh, Epps, and Bailey]{vinh2009information}
Nguyen~Xuan Vinh, Julien Epps, and James Bailey.
\newblock Information theoretic measures for clusterings comparison: is a correction for chance necessary?
\newblock In \emph{Proceedings of the 26th annual international conference on machine learning}, pp.\  1073--1080, 2009.

\bibitem[Wang et~al.(2022)Wang, Shan, Zhang, and Yang]{wang2022n24news}
Zhen Wang, Xu~Shan, Xiangxie Zhang, and Jie Yang.
\newblock N24news: A new dataset for multimodal news classification.
\newblock In \emph{13th International Conference on Language Resources and Evaluation Conference, LREC 2022}, pp.\  6768--6775. European Language Resources Association (ELRA), 2022.

\bibitem[Watanabe(1960)]{watanabe1960information}
Satosi Watanabe.
\newblock Information theoretical analysis of multivariate correlation.
\newblock \emph{IBM Journal of research and development}, 4\penalty0 (1):\penalty0 66--82, 1960.

\bibitem[Wold et~al.(1987)Wold, Esbensen, and Geladi]{wold1987principal}
Svante Wold, Kim Esbensen, and Paul Geladi.
\newblock Principal component analysis.
\newblock \emph{Chemometrics and intelligent laboratory systems}, 2\penalty0 (1-3):\penalty0 37--52, 1987.

\bibitem[Wolf et~al.(2020)Wolf, Debut, Sanh, Chaumond, Delangue, Moi, Cistac, Rault, Louf, Funtowicz, et~al.]{wolf2020transformers}
Thomas Wolf, Lysandre Debut, Victor Sanh, Julien Chaumond, Clement Delangue, Anthony Moi, Pierric Cistac, Tim Rault, R{\'e}mi Louf, Morgan Funtowicz, et~al.
\newblock Transformers: State-of-the-art natural language processing.
\newblock In \emph{Proceedings of the 2020 conference on empirical methods in natural language processing: system demonstrations}, pp.\  38--45, 2020.

\bibitem[Young et~al.(2014)Young, Lai, Hodosh, and Hockenmaier]{young2014image}
Peter Young, Alice Lai, Micah Hodosh, and Julia Hockenmaier.
\newblock From image descriptions to visual denotations: New similarity metrics for semantic inference over event descriptions.
\newblock \emph{Transactions of the Association for Computational Linguistics}, 2:\penalty0 67--78, 2014.

\bibitem[Zhai et~al.(2023)Zhai, Mustafa, Kolesnikov, and Beyer]{zhai2023sigmoid}
Xiaohua Zhai, Basil Mustafa, Alexander Kolesnikov, and Lucas Beyer.
\newblock Sigmoid loss for language image pre-training.
\newblock In \emph{Proceedings of the IEEE/CVF International Conference on Computer Vision}, pp.\  11975--11986, 2023.

\bibitem[Zhen et~al.(2022)Zhen, Meng, Chakraborty, and Singh]{zhen2022versatile}
Xingjian Zhen, Zihang Meng, Rudrasis Chakraborty, and Vikas Singh.
\newblock On the versatile uses of partial distance correlation in deep learning.
\newblock In \emph{Proceedings of the European conference on computer vision (ECCV)}, 2022.

\end{thebibliography}
